\documentclass{article}

\PassOptionsToPackage{numbers, compress}{natbib}
\usepackage{amsmath}
\usepackage{amssymb}
\usepackage{mathtools}
\usepackage{amsthm}
\usepackage[most]{tcolorbox}
\usepackage{fvextra}
\usepackage[T1]{fontenc}

\usepackage{microtype}
\usepackage{graphicx}
\usepackage{subcaption}
\usepackage{booktabs} % for professional tables
\usepackage{multirow}
\usepackage[hidelinks]{hyperref}

\usepackage{natbib}
\usepackage{amsmath}
\usepackage{amssymb}
\usepackage{mathtools}
\usepackage{amsthm}
\usepackage[most]{tcolorbox}
\usepackage{fvextra}
\usepackage{algorithm}
\usepackage{algorithmic}

\usepackage[capitalize,noabbrev]{cleveref}

\usepackage{silence}
\newcommand{\method}{{ATHENA}}

\newtcolorbox{promptbox}{
  enhanced,
  breakable,
  colback=gray!3,
  colframe=gray!50,
  boxrule=0.5pt,
  arc=2mm,
  left=6pt,
  right=6pt,
  top=6pt,
  bottom=6pt,
  before skip=10pt,
  after skip=10pt,
  listing only,
}

\newtcolorbox{CodeFence}{
  enhanced,
  breakable,
  colback=black!3,
  colframe=black!0,   % no visible frame
  boxrule=0pt,
  arc=1mm,
  left=8pt, right=6pt, top=6pt, bottom=6pt,
  borderline west={2pt}{0pt}{black!30}, % left vertical bar like code block
  before skip=6pt,
  after skip=6pt,
}

\theoremstyle{plain}

\theoremstyle{definition}

\theoremstyle{remark}

\usepackage{arxiv}

\title{ATHENA: Adaptive Test-Time Steering for \\ Improving Count Fidelity in Diffusion Models}

\author{
\textbf{Mohammad Shahab Sepehri}\thanks{Equal contribution.} \quad
Asal Mehradfar\footnotemark[1] \quad
Berk Tinaz \quad
Salman Avestimehr \quad
Mahdi Soltanolkotabi \\
Department of Electrical and Computer Engineering \\
University of Southern California, Los Angeles, CA, USA \\
\texttt{\{sepehri, mehradfa, tinaz, avestime, soltanol\}@usc.edu}
}

\begin{document}

\maketitle

\begin{abstract}
Text-to-image diffusion models achieve high visual fidelity but surprisingly exhibit systematic failures in numerical control when prompts specify explicit object counts. To address this limitation, we introduce \method{}, a model-agnostic, test-time adaptive steering framework that improves object count fidelity without modifying model architectures or requiring retraining. \method{} leverages intermediate representations during sampling to estimate object counts and applies count-aware noise corrections early in the denoising process, steering the generation trajectory before structural errors become difficult to revise. We present three progressively more advanced variants of \method{} that trade additional computation for improved numerical accuracy, ranging from static prompt-based steering to dynamically adjusted count-aware control. Experiments on established benchmarks and a new visually and semantically complex dataset show that \method{} consistently improves count fidelity, particularly at higher target counts, while maintaining favorable accuracy--runtime trade-offs across multiple diffusion backbones. Our code and data are publicly available at \url{https://github.com/MShahabSepehri/ATHENA}.
\end{abstract}

\section{Introduction}

% Text-to-image (T2I) diffusion models have become the dominant paradigm for high-quality image generation, achieving strong visual realism, semantic alignment, and compositional expressiveness~\cite{rombach2022high}. 
% As these models are increasingly deployed as controllable components within larger systems rather than standalone creative tools, their ability to reliably satisfy structured and quantitative constraints has emerged as a critical practical requirement.
Text-to-image (T2I) diffusion models have gained huge popularity and become the dominant paradigm for high-quality image generation, driven by their ability to synthesize images from natural language prompts with strong visual realism, semantic alignment, and compositional expressiveness~\cite{rombach2022high}.

\begin{figure}[t]
    \centering
    \includegraphics[width=0.99\linewidth]{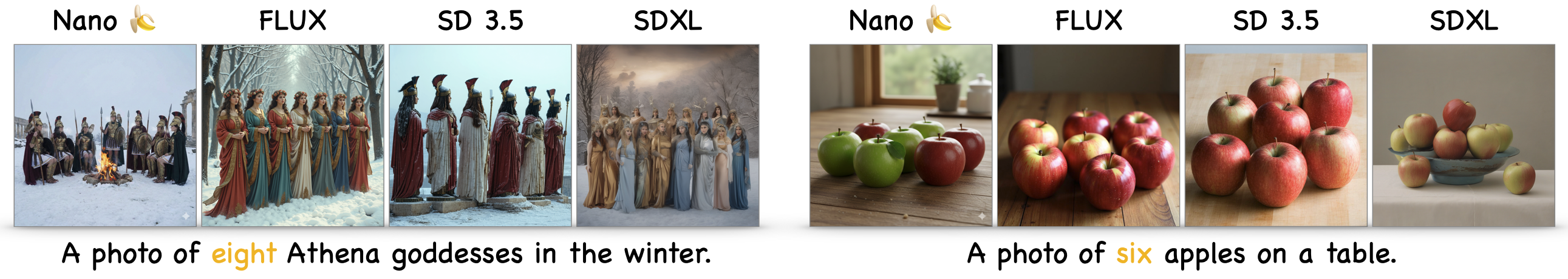}
    \caption{Count fidelity failures in T2I generative models. Despite explicit numerical specifications in the prompts, generated images frequently contain incorrect object counts across multiple generative backbones, even for simple objects and scenes, highlighting a persistent limitation in numerical control.}
    \label{fig:counting_failures}
\end{figure}

Despite their advances in generating high-quality images with fine details, diffusion models continue to exhibit systematic failures in numerical control when prompts specify explicit object counts. This limitation is particularly striking given the remarkable generative capabilities of modern diffusion models, which can synthesize complex scenes with rich semantics and fine-grained details. Even for visually simple and well-defined categories, prompts such as \emph{``a photo of eight cows in a field''} or \emph{``ten apples on a table''} often produce images with missing instances, duplicated objects, or ambiguous merges that obscure instance boundaries.
While prompt sensitivity can influence generation outcomes and has motivated prompt optimization strategies~\cite{gan2025conceptmixpp}, such interventions remain insufficient for reliably enforcing quantitative constraints~\cite{ramesh2022hierarchical}. As illustrated in \cref{fig:counting_failures}, these counting errors persist across multiple state-of-the-art T2I generative models, indicating a general limitation rather than a model-specific shortcoming.

Accurate object counting is important beyond visual plausibility. Downstream applications such as synthetic data generation for object detection and segmentation~\cite{nikolenko2021synthetic,tremblay2018training} and simulation environments for robotics and embodied agents~\cite{makoviychuk2021isaac} rely on faithful instance-level control. In these settings, counting errors can propagate to subsequent perception or decision-making stages, undermining the reliability of diffusion-based generators when used as components within larger systems.

From a modeling perspective, object counting poses a natural challenge for diffusion-based generation. Diffusion models rely on a continuous denoising process without explicitly representing discrete object instances or enforcing global numerical constraints~\cite{ho2020denoising,ho2022classifier}. Early stochastic decisions in the sampling trajectory largely determine object multiplicity and spatial layout, while later steps primarily refine local appearance~\cite{li2025enhancing, tinaz2025emergence, mahajan2024prompting, hertz2023prompttoprompt, patashnik2023localizing}. As a result, counting errors are difficult to correct once they emerge, as object count is tightly entangled with occlusion, spatial arrangement, and semantic diversity~\cite{lee2025countcluster}. These characteristics motivate intervention mechanisms that act before global structural decisions become effectively irreversible.

Several recent works have explored mechanisms for improving counting fidelity in diffusion models, including guidance-based and iterative refinement approaches~\cite{binyamin2025makeitcount,zafar2024iterative}. While these methods can improve counting accuracy under controlled settings, they often rely on model-specific assumptions or lack adaptive mechanisms to correct errors once the sampling trajectory deviates from the target count. As a result, achieving reliable and efficient count control that generalizes across diffusion backbones and realistic prompt distributions remains an open challenge.

In this work, we introduce \method{} (Adaptive Trajectory Harmonization via Early Numerical Assessment), a test-time \emph{adaptive} steering framework for improving object count fidelity in T2I diffusion models. \method{} estimates object count from intermediate representations and uses this signal to adaptively modify the sampling trajectory via count-aware noise correction. By intervening early in the denoising process, the method enables corrective control before key structural 
layouts are formed~\cite{karras2022elucidating,choi2022perception}. \method{} is model-agnostic, requires no architectural modifications or retraining, and can be applied across diverse diffusion backbones.

We instantiate \method{} through a progression of steering strategies that isolate the role of adaptivity in count control. \method{}-Static applies a fixed steering signal, \method{}-Feedback introduces conditional, count-aware steering based on intermediate estimates, and \method{}-Adaptive, our primary method, further adjusts steering strength online in response to persistent count errors. We evaluate these variants on three datasets: the CoCoCount benchmark~\cite{binyamin2025makeitcount}, a balanced extension spanning target counts from two to ten (CoCoCount-E), and a newly constructed challenging dataset (\method{} dataset). Across datasets and diffusion backbones, our adaptive test-time steering improves count fidelity while maintaining favorable accuracy--runtime trade-offs.

\paragraph{Contributions.}
Our main contributions are as follows:
\begin{itemize}
    % \item We propose \method{}, a \emph{model-agnostic} test-time adaptive steering framework for improving object count fidelity in T2I diffusion models, without architectural modifications or retraining.

    % \item We introduce three instantiations of \method{}, namely \method{}-Static, \method{}-Feedback, and \method{}-Adaptive, which progressively incorporate conditional and adaptive steering, isolating the role of adaptivity in count control.

    % \item We construct a new evaluation dataset (\method{} dataset) that complements existing counting benchmarks by targeting challenging object categories and compositional prompt structures.

    % \item We demonstrate across multiple diffusion backbones and datasets that adaptive test-time steering improves count fidelity while maintaining favorable accuracy--runtime trade-offs.

    \item We propose \method{}, a model-agnostic steering framework for improving object-count fidelity in T2I diffusion models that requires no architectural changes or retraining. \method{} can be applied to arbitrary T2I models via simple forward-pass interventions and yields immediate improvements in numerical fidelity.
    % and enable a controlled study of the role of adaptivity in numerical control.

    \item We introduce the \method{} dataset, a new evaluation benchmark that complements existing counting datasets by targeting challenging object categories (e.g., 
    % snow globe, 
    accordion) and compositional prompt structures with relational constraints (e.g., next to a river) and object-level distractions (e.g., with a person).

    % \item Through extensive experiments on three datasets, we demonstrate that \method{} improves the numerical accuracy of the base diffusion models by up to \textbf{22\%} and outperforms baselines, while reducing memory usage by approximately \textbf{4$\times$} and achieving up to \textbf{2.5$\times$} faster image generation relative to the baselines.
    \item Through extensive experiments on three datasets, we demonstrate that \method{} improves the numerical accuracy of the base diffusion models by up to 22\% and outperforms baselines, while reducing memory usage by approximately $4\times$ and achieving up to $2.5\times$ faster image generation relative to the baselines.

    % \item Through extensive experiments on three datasets, we demonstrate that \method{} improves the numerical accuracy of the original diffusion models by up to 22 percentage points and outperforms baselines, while reducing memory usage by approximately $4\times$ and achieving up to $2.5\times$ faster image generation relative to the baselines.
\end{itemize}
\section{Related Work}

Prior work on numerical control in T2I diffusion models has identified object counting as a persistent challenge when prompts specify explicit cardinalities~\cite{wu2024conceptmix, gan2025conceptmixpp}.
\emph{Make It Count}~\cite{binyamin2025makeitcount} improves counting by training an auxiliary layout-modification model that leverages SDXL-specific internal features to detect counting discrepancies and adjust object layouts, but its reliance on additional training and model-specific internals limits generality and lightweight test-time deployment.
In contrast, \emph{CountCluster}~\cite{lee2025countcluster} enforces quantity control by encouraging early cross-attention activations to form a fixed number of spatial clusters matching the target object count, implicitly assuming uncluttered scenes with well-separated instances and degrading in the presence of overlap or complex background structure.

% \begin{figure*}[t]
%     \centering
%     \includegraphics[width=0.92\linewidth]{figures/diffusion_pipeline.png}
%     \caption{Count estimation across diffusion steps. At early sampling steps, object structure is weakly formed and decoded images are too noisy for reliable count estimation. At later steps, object instances become well defined, and counting is reliable, but requires additional diffusion steps and a higher inference cost. This trade-off motivates estimating object count at an intermediate diffusion step.
%     }
%     \label{fig:count_observability}
% \end{figure*}

Several methods regulate object counts by introducing gradient-based guidance or optimization during diffusion sampling.
\emph{Counting Guidance}~\cite{kang2025counting} enforces target counts via classifier guidance, using gradients from a trained counting network at each denoising step, which increases inference cost and can disrupt global structure when large corrections are applied.
Similarly, \emph{YOLO-Count}~\cite{zeng2025yolo} applies classifier-style gradient guidance based on dense cardinality predictions from a specialized counting model, but requires additional training and does not adaptively respond to persistent count errors during sampling.
In contrast, \emph{Detection-Driven Object Count Optimization}~\cite{zafar2024iterative} corrects count errors through iterative test-time optimization using feedback from an external object detector, but relies on repeated evaluation and late-stage refinement, leading to slower inference and limited effectiveness once early structural decisions are fixed.

Training-free and pipeline-based alternatives have also been explored to address counting errors without modifying diffusion models.
\emph{CountDiffusion}~\cite{li2025countdiffusion} modifies attention maps based on external count estimates from intermediate generations, but relies on task-specific auxiliary counters rather than adaptive steering.
Beyond diffusion-level intervention, decomposition- or agent-based pipelines generate objects sequentially or via sub-task planning (e.g., \emph{MuLan}~\cite{li2024mulan}), but introduce computational overhead and deviate from standard diffusion sampling, limiting scalability and practical deployment.

More broadly, guidance and steering techniques have been widely studied in diffusion models to improve conditional alignment.
Attention-based interventions manipulate cross-attention to enhance semantic or spatial consistency~\cite{hertz2023prompttoprompt,chefer2023attend}, while reward-based approaches refine image--text alignment using external signals~\cite{bahng2025cycle}.
However, these methods primarily target semantic fidelity or attribute binding and do not enforce discrete global constraints such as exact object counts, particularly at higher cardinalities where early stochastic decisions dominate.

External perception models, such as object detectors and vision--language models, have been used to guide or evaluate diffusion-based generation~\cite{liu2023detector,li2022grounded,liu2024grounding,cheng2024yolo, sepehri2025hyperphantasia}.
Prior work typically employs these models as controllers or optimization objectives, coupling generation quality to detector behavior and increasing inference cost.
In contrast, \method{} uses external models only for intermediate count estimation and leverages this signal to adaptively steer the diffusion trajectory, without architectural changes or iterative optimization.

Overall, existing approaches to improving object count fidelity rely on model-specific internals, incur substantial computational overhead, or compromise visual quality. Efficient, adaptive, and model-agnostic count control at test time remains an open challenge, which \method{} addresses.
\section{Preliminaries}

\subsection{Diffusion Models for Text-to-Image Generation}

T2I diffusion models synthesize images by iteratively transforming noise into data through a sequence of denoising steps.
At sampling step $t$, a learned denoiser $\epsilon_\theta(\cdot)$ is applied to the current latent representation $z_t$ conditioned on a text prompt $p$. We denote the denoiser output by $\epsilon_t \triangleq \epsilon_\theta(t, z_t, p)$. The latent state is then updated using a sampler-specific transition operator,
\begin{equation}
z_{t-1} = \mathcal{S}_t(z_t, \epsilon_t, p),
\label{eqn:sampler}
\end{equation}
where $\mathcal{S}_t(\cdot)$ may be stochastic or deterministic depending on the sampling scheme, encompassing diffusion samplers such as DDPM~\cite{ho2020denoising} and deterministic probability-flow variants~\cite{song2021scorebased} used in modern systems.

The denoiser output also defines an estimate of the underlying clean latent at step $t$, denoted by $\hat z_0^{(t)}$, via a sampler-defined reconstruction operator
\begin{equation}
\hat z_0^{(t)} = \mathcal{D}_t(z_t, \epsilon_t).
\end{equation}
For the diffusion models evaluated in this work, this reconstruction is given by
\begin{equation}
\hat z_0^{(t)} = z_t - \sigma_t \epsilon_t,
\end{equation}
where $\sigma_t$ is a scheduler-dependent noise scale.
The estimated clean latent may be decoded into image space for intermediate analysis when required.

\subsection{Conditional Guidance and Sampling Dynamics}\label{sec:guidance}

\begin{figure*}[t]
    \centering
    \includegraphics[width=0.92\linewidth]{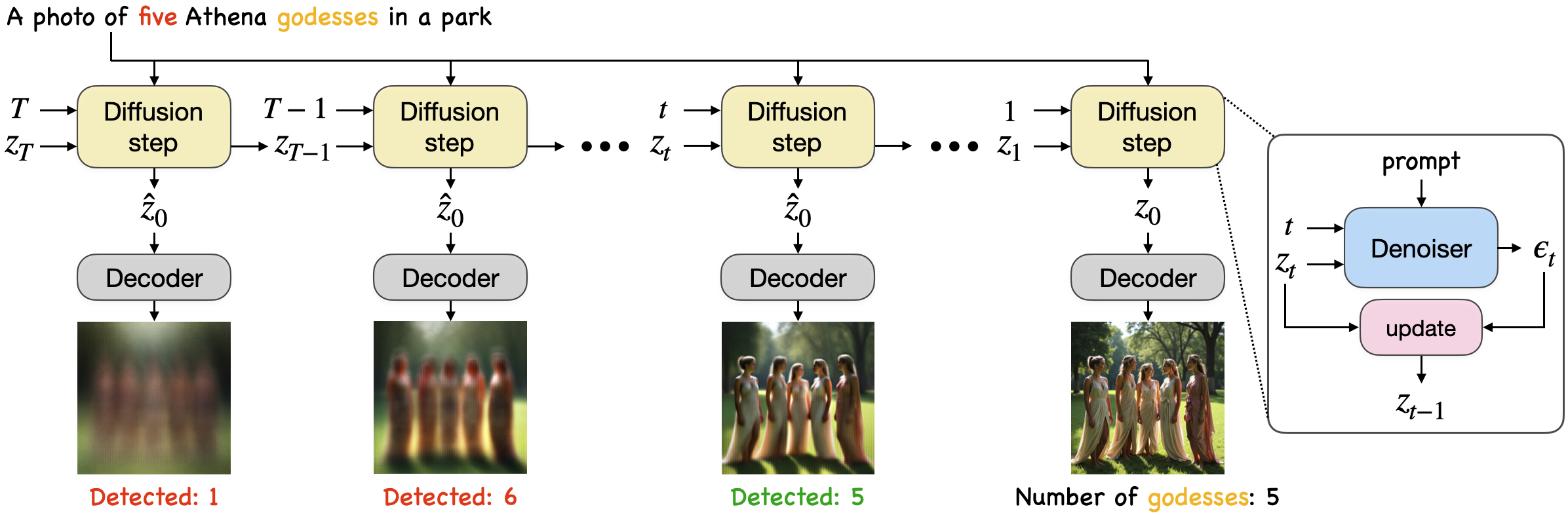}
    \caption{Count estimation across diffusion steps. At early sampling steps, object structure is weakly formed and decoded images are too noisy for reliable count estimation. At later steps, object instances become well defined, and counting is reliable, but requires additional diffusion steps and a higher inference cost. This trade-off motivates estimating object count at an intermediate diffusion step.
    }
    \label{fig:count_observability}
\end{figure*}

Conditional control in diffusion-based T2I models is commonly achieved via guidance, where the sampling trajectory is modified to satisfy a desired objective.
Given a conditioning loss $\mathcal{L}(z_t, p)$, guidance-based methods adjust the denoiser output as
\begin{equation}
\hat{\epsilon}_t = \epsilon_t - s \, \nabla_{z_t} \mathcal{L}(z_t, p),
\end{equation}
where $\epsilon_t \triangleq \epsilon_\theta(t, z_t, p)$ and $s$ controls the guidance strength~\cite{dhariwal2021diffusion,ho2022classifier}.

Such guidance is effective for semantic or appearance-level control but is less suited for global structural properties, such as object count, which are largely determined during early sampling steps. This limitation motivates adaptive interventions that operate early in the sampling trajectory, before global structure becomes fixed.
\section{Method}
\subsection{Problem Setup and Design Goals}

We study test-time control of object count fidelity in diffusion-based T2I models.
Let $G_\theta$ denote a pretrained generator that produces an image through an iterative sampling process, yielding a sequence of latent representations $\{z_t\}_{t=0}^T$ conditioned on a text prompt $p$ specifying a target object count $k$ for a particular object category.
Our objective is to improve the accuracy with which the final generated image satisfies this target count.
We assume access to intermediate latent or decoded representations during sampling, and all interventions are performed at test time without retraining or modifying the generator parameters.

\method{} frames object counting as a discrete control problem over the diffusion sampling trajectory.
It operates by monitoring intermediate generation signals and applying lightweight, test-time adjustments that influence the trajectory.
These design goals define the class of adaptive, test-time interventions considered in the following sections.

\subsection{Count Estimation Across Diffusion Steps}
\label{sec:count_est}

Diffusion-based T2I generation proceeds through a sequence of intermediate latent representations that transition from noise to a coherent image.
At early sampling steps, object structure is insufficiently formed and decoded images are too noisy for reliable automated count estimation.
At later steps, object instances are fully formed and count estimation becomes reliable, but this requires additional diffusion steps, increasing inference time. As illustrated in \cref{fig:count_observability}, this tradeoff yields an intermediate regime in which object count can be estimated reliably without incurring the cost of late-stage sampling.

Motivated by this observation, \method{} performs count estimation at a fixed intermediate diffusion step $t_{\text{est}}$, decoding the corresponding clean latent estimate $\hat z_0^{(t_{\text{est}})}$ to balance estimation reliability and inference cost.

\subsection{\method{}: Test-Time Steering Framework}
\label{sec:method}

\method{} is a test-time steering framework for improving object count fidelity in diffusion-based T2I generation, without modifying model parameters or requiring retraining.
The framework estimates object count at an intermediate diffusion step and uses this signal to steer the sampling trajectory via prompt-based control, enabling corrective intervention before structural errors form.
All components operate at inference time and are compatible with both deterministic and stochastic diffusion samplers.

\subsubsection{Prompt-Based Steering Mechanism}\label{sec:prompt_control}

% At the core of \method{} is a lightweight, training-free steering mechanism that modifies the denoising trajectory via prompt conditioning.
% At diffusion step $t$, the denoiser is evaluated twice on the current latent $z_t$: once with the original prompt $p$, producing
% $\epsilon_t \triangleq \epsilon_\theta(t, z_t,p)$, and once with a \emph{control prompt} $\hat p$, producing
% $\hat\epsilon_t \triangleq \epsilon_\theta(t, z_t,\hat p)$ (see \cref{fig:athena_block}).
% These two predictions are combined to form a steered noise estimate
% \begin{equation}
% \tilde\epsilon_t \;=\; \epsilon_t + \gamma\,(\epsilon_t - \hat\epsilon_t),
% \label{eq:athena_steer}
% \end{equation}
% where $\gamma \ge 0$ controls the steering strength.
% The sampler update (\cref{eqn:sampler}) is then applied using $\tilde\epsilon_t$ in place of $\epsilon_t$ to obtain $z_{t-1}$. We normalize $\tilde\epsilon_t$ to match the norm of $\epsilon_t$, ensuring the steered latent remains within the denoiser’s expected scale and preserves stable, in-distribution sampling.

% \begin{figure}[h]
%     \centering
%     \includegraphics[width=0.4\linewidth]{figures/ATHENA_step.png}
%     \caption{\method{} steering block. The denoiser is evaluated with the original and control prompts to obtain $\epsilon_t$ and $\hat{\epsilon}_t$, which are combined to obtain the latent state $z_{t-1}$ for test-time count control.
%     }
%     \label{fig:athena_block}
% \end{figure}

\begin{figure}[h]
\centering
\begin{minipage}[ht]{0.57\linewidth}
    \vspace{0pt}
    At the core of \method{} is a lightweight, training-free steering mechanism that modifies the denoising trajectory via prompt conditioning.
    At diffusion step $t$, the denoiser is evaluated twice on the current latent $z_t$: once with the original prompt $p$, producing
    $\epsilon_t \triangleq \epsilon_\theta(t, z_t,p)$, and once with a \emph{control prompt} $\hat p$, producing
    $\hat\epsilon_t \triangleq \epsilon_\theta(t, z_t,\hat p)$ (see \cref{fig:athena_block}).
    These two predictions are combined to form a steered noise estimate
    \begin{equation}
    \tilde\epsilon_t \;=\; \epsilon_t + \gamma\,(\epsilon_t - \hat\epsilon_t),
    \label{eq:athena_steer}
    \end{equation}
    where $\gamma \ge 0$ controls the steering strength.
    The sampler update (\cref{eqn:sampler}) is then applied using $\tilde\epsilon_t$ in place of $\epsilon_t$ to obtain $z_{t-1}$. We normalize $\tilde\epsilon_t$ to match the norm of $\epsilon_t$, ensuring the steered latent remains within the denoiser’s expected scale and preserves stable, in-distribution sampling.
\end{minipage}
\hfill
\begin{minipage}[ht]{0.41\linewidth}
    \vspace{0pt}
    \centering
    \includegraphics[width=\linewidth]{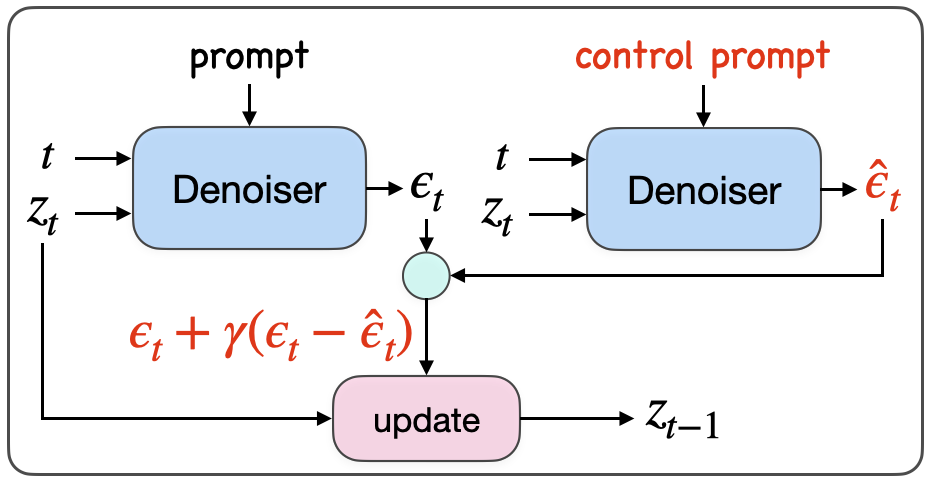}
    \caption{\method{} steering block. The denoiser is evaluated with the original and control prompts to obtain $\epsilon_t$ and $\hat{\epsilon}_t$, which are combined to obtain the latent state $z_{t-1}$ for test-time count control.
    }
    \label{fig:athena_block}
\end{minipage}
\end{figure}

The steering operation in \eqref{eq:athena_steer} can be interpreted as a controlled perturbation of the denoiser’s prediction that biases the sampling trajectory away from behaviors induced by the control prompt.
The difference term $(\epsilon_t - \hat\epsilon_t)$ isolates the directional change in predicted noise when replacing the original prompt $p$ with $\hat p$, capturing the local influence of the control condition at step $t$.
By adding a scaled version of this difference to $\epsilon_t$, \method{} reinforces directions that move the generation toward the target count while preserving semantic structure.
    
From a dynamical perspective, this update modifies the effective vector field of the reverse diffusion process, inducing a prompt-dependent drift without altering model parameters or performing gradient-based optimization.
Unlike classifier- or loss-based guidance (\cref{sec:guidance}), the steering term is computed through forward denoiser evaluations, making it compatible with both stochastic and deterministic samplers.
The scalar $\gamma$ controls the magnitude of this perturbation, enabling smooth interpolation between unsteered sampling ($\gamma=0$) and stronger corrective influence.

\subsubsection{Control Prompt Construction}

\method{} induces steering by evaluating the denoiser under two textual conditionings: the original prompt $p$ and a modified \emph{control prompt} $\hat p$, whose effect on the denoiser output is formalized in \cref{eq:athena_steer}.
The control prompt provides an alternative conditioning signal that encodes corrective intent purely at the prompt level, while remaining fully compatible with the pretrained generator.

The framework imposes minimal assumptions on the form of $\hat p$.
In practice, $\hat p$ is constructed by modifying only the object-count specification in the original prompt $p$, while keeping all other semantic content unchanged.
We consider two forms of control prompts: (i) a \emph{count-agnostic} prompt obtained by removing explicit cardinality constraints, and (ii) a \emph{feedback-based} prompt in which the target count is replaced by an estimated count from an intermediate decoded generation.
In all cases, $\hat p$ is treated identically to $p$ by the underlying model and requires no architectural changes, auxiliary networks, or gradient-based optimization.

By separating control prompt construction from the steering mechanism, \method{} decouples how corrective information is encoded from how it influences the sampling trajectory.
Different choices of $\hat p$ yield distinct instantiations of the framework, while sharing the same denoiser evaluations and sampler interaction.
We describe these instantiations in the following subsection.

\subsubsection{\method{} Control Strategies}
\label{sec:strategies}

We present three instantiations of \method{} that progressively increase adaptivity while preserving the same test-time, training-free steering mechanism. All strategies rely on the prompt-based steering operation described in \cref{sec:prompt_control} and differ in whether and how intermediate generation signals are used to select the control prompt and steering strength. This progression moves from fixed, count-agnostic control to feedback-informed and adaptive steering, enabling increasingly targeted correction of counting errors while maintaining a lightweight, model-agnostic design.

\paragraph{\method{}-Static.}
\label{sec:athena_static}

\method{}-Static applies prompt-based steering using a fixed, count-agnostic control prompt.
Given an original prompt $p$ specifying a target count $k$, the control prompt $\hat p$ is constructed by removing the explicit cardinality constraint while preserving all other semantic content.
% Given an original prompt $p$ specifying a target count $k$ (e.g., ``a photo of seven apples''), the control prompt $\hat{p}$ is constructed by removing the explicit cardinality constraint while preserving all other semantic content (e.g., ``a photo of apples'').
No intermediate count estimation is performed, and steering is applied once from the initial diffusion step to a cutoff $t_{\mathrm{steer}}$, after which standard diffusion proceeds unmodified.

This variant isolates the effect of prompt-level steering without feedback or adaptive adjustment.
It incurs minimal computational overhead and serves as a baseline demonstrating that count fidelity can be influenced through prompt-based steering.
Details are provided in \cref{appx:static}.

\paragraph{\method{}-Feedback.}
\label{sec:athena_feedback}

While static steering can improve count fidelity, its effectiveness depends on how well a fixed control prompt aligns with the trajectory.
\method{}-Feedback addresses this limitation by incorporating a single intermediate count estimate to inform the control prompt.

Specifically, diffusion is first run without steering until an intermediate step $t_{\mathrm{est}}$, where the partially denoised latent is decoded and the object count is estimated.
If the estimated count differs from the target, generation is restarted from the same initial noise using a feedback control prompt constructed by replacing the original count in $p$ with the observed count.
Prompt-based steering is then applied once during early diffusion steps down to a cutoff $t_{\mathrm{steer}}$, after which diffusion proceeds unmodified to completion.
No additional count checks or parameter updates are performed.

By correcting semantic mismatch between the prompt and the emerging structure, \method{}-Feedback improves robustness over static steering while remaining simple and efficient.
Algorithmic details are provided in \cref{appx:feedback}.

\paragraph{\method{}-Adaptive.}
\label{sec:athena_adaptive}

Although feedback steering makes the control prompt feedback-informed, its success depends on the steering strength $\gamma$.
If $\gamma$ is too small, steering may reduce the counting error without reaching the target; if $\gamma$ is too large, it can overshoot the target and flip the error direction, often at the expense of structural quality.
\method{}-Adaptive resolves this sensitivity through a single, direction-aware adjustment of $\gamma$ based on intermediate feedback.

\begin{figure}[ht]
    \centering
    \includegraphics[width=0.55\linewidth]{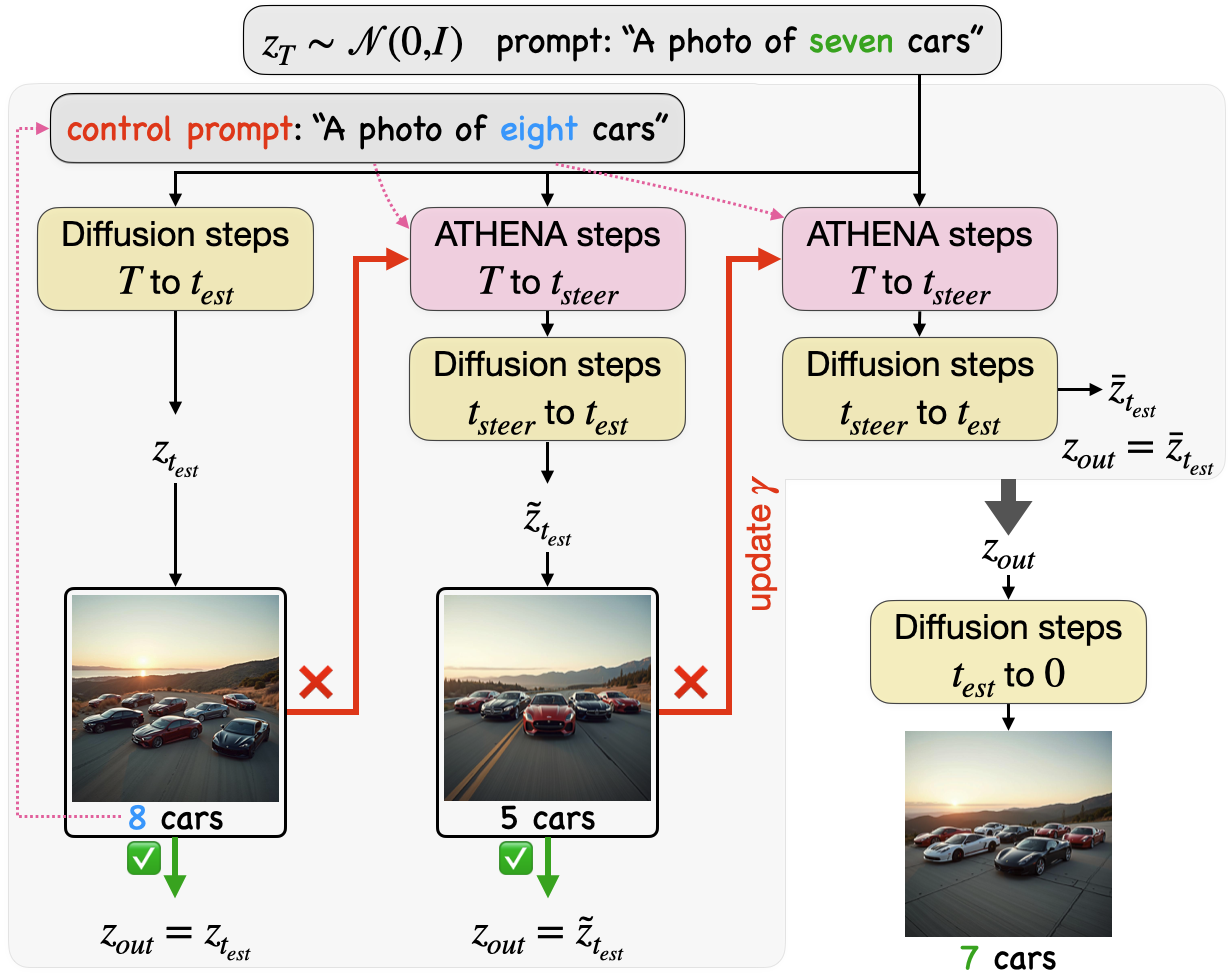}
    \caption{\method{}-Adaptive pipeline. The method estimates object count at an intermediate diffusion step, applies early-stage prompt-based steering, adaptively adjusts the steering strength once based on the observed error direction, and completes generation using standard diffusion.
    }
    \label{fig:athena_adaptive}
\end{figure}

As illustrated in \cref{fig:athena_adaptive}, the method first estimates the baseline count at $t_{\mathrm{est}}$ without steering.
If the estimate deviates from the target, prompt-based steering is applied once using the initial $\gamma$.
If the resulting count still deviates from the target, the observed change indicates whether steering moved the generation toward or away from the target.
When the error direction remains unchanged, $\gamma$ is doubled; when it flips, $\gamma$ is halved.
A final steered generation is then performed using the adjusted $\gamma$, after which diffusion proceeds unmodified to completion.

Importantly, \method{}-Adaptive does not perform iterative optimization or repeated parameter tuning.
The error direction observed after a single feedback step provides a signal for adjusting the steering magnitude, striking a balance between corrective strength and structural stability.
Further adjustment is unnecessary in practice, as additional iterations increase computational cost with diminishing returns.
As a result, the method requires at most two steered trajectories, introduces no additional learned components, and improves robustness while preserving favorable accuracy--runtime trade-offs.
Pseudocode is provided in \cref{appx:adaptive}.
\section{Experiments}

We evaluate \method{} across multiple T2I diffusion backbones and counting benchmarks to assess effectiveness, robustness, and computational trade-offs.
Our experiments span three datasets of increasing complexity, including a newly introduced \method{} dataset, and compare static, feedback, and adaptive steering against existing baselines.
We report quantitative accuracy, analyze performance as a function of target count, and examine accuracy--runtime trade-offs alongside qualitative examples illustrating the impact of adaptive test-time steering.

\subsection{Experimental Setup}

\textbf{Models.}
We evaluate \method{} on three representative T2I diffusion backbones spanning different architectures and training regimes: SDXL~\cite{SDXL, podell2023sdxl}, SD~3.5 Large~\cite{SD3.5large, esser2024scaling}, and FLUX.1-dev~\cite{flux2024, labs2025flux}.
All models are used as released, without modification or retraining.

\textbf{Datasets.}
We conduct experiments on three counting benchmarks of increasing complexity.
\emph{CoCoCount}~\cite{binyamin2025makeitcount} consists of simple prompts with explicit numerical constraints and minimal scene context.
To avoid bias from repeated prompts in the original benchmark, we use a deduplicated version of the dataset in which prompts with identical text are retained once, yielding 161 unique prompts with target counts from 2 to 10 (see \cref{appx:datasets}).
While well-suited for controlled evaluation, CoCoCount offers limited diversity in prompt structure and scene complexity.

To enable systematic analysis across target cardinalities, we construct \emph{CoCoCount-E}, an extended benchmark in which a filtered subset of CoCoCount prompts is instantiated with target counts from 2 to 10, yielding 648 distinct prompts with consistent structure and a balanced count distribution.
We cap the maximum target count at 10, as prior work~\cite{binyamin2025makeitcount} reports substantial degradation beyond this range, which we also observe empirically.

We further introduce the \method{} dataset, a new benchmark designed to evaluate object counting under progressively challenging prompt conditions.
The dataset consists of 360 prompts organized into four levels of increasing complexity, with target counts ranging from 2 to 10 (see \cref{tab:athena_dataset}).
Prompts are generated using a large language model to systematically combine multiple constraints within a single instruction, ensuring controlled diversity across difficulty levels. Additional construction details and statistics for all benchmarks are provided in \cref{appx:datasets}.

\begin{table*}[h]
\centering
\caption{Structure of the \method{} dataset with four levels of increasing prompt complexity.}
\label{tab:athena_dataset}
\small
% \resizebox{0.7\textwidth}{!}{
\begin{tabular}{lll}
\toprule
\textbf{Level} & \textbf{Description} & \textbf{Example Prompt} \\
\midrule
L1 & Hard object categories & ``nine microphones'' \\
L2 & + Scene context & ``eight sneakers at the edge of a pond'' \\
L3 & + Distractor objects & ``five jars beside a river with a person'' \\
L4 & + Relational constraints & ``seven dumbbells lined up along a sidewalk'' \\
\bottomrule
\end{tabular}
% }
\end{table*}

\begin{table*}[t]
\centering
\caption{Quantitative counting performance across diffusion backbones and datasets.
Accuracy (\%) is reported as exact-match count accuracy, with the best result for each model--dataset pair shown in \textbf{bold}.
MAE and RMSE measure counting error, and Time denotes mean generation time per sample (seconds).
}
\label{tab:quant_results}
\small
\setlength{\tabcolsep}{5pt}
\resizebox{0.99\textwidth}{!}{
\begin{tabular}{l l cccc cccc cccc}
\toprule
\multirow{2}{*}{\textbf{Model}} &
\multirow{2}{*}{\textbf{Method}} &
\multicolumn{4}{c}{\textbf{CoCoCount}} &
\multicolumn{4}{c}{\textbf{CoCoCount-E}} &
\multicolumn{4}{c}{\textbf{\method{} Dataset}} \\
\cmidrule(lr){3-6} \cmidrule(lr){7-10} \cmidrule(lr){11-14}
& & 
Acc~($\uparrow$) & MAE~($\downarrow$) & RMSE~($\downarrow$) & Time~($\downarrow$) &
Acc~($\uparrow$) & MAE~($\downarrow$) & RMSE~($\downarrow$) & Time~($\downarrow$) &
Acc~($\uparrow$) & MAE~($\downarrow$) & RMSE~($\downarrow$) & Time~($\downarrow$) \\
\midrule

\multirow{4}{*}{\textbf{FLUX.1-dev}}
& Unsteered        & 58.4 & 0.98 & 1.96 & 45.8 & 46.5 & 1.12 & 1.93 & 22.6 & 39.4 & 1.78 & 2.96 & 45.5 \\
& \method{}-Static & \textbf{73.3} & 0.56 & 1.35 & 54.8 & 58.5 & 0.84 & 1.73 & 27.4 & 48.9 & 1.31 & 2.44 & 54.7 \\
& \method{}-Feedback & 71.4 & 0.67 & 1.72 & 56.0 & 58.3 & 0.98 & 2.00 & 29.8 & 52.2 & 1.38 & 2.60 & 60.8 \\
& \method{}-Adaptive & 70.2 & 0.65 & 1.63 & 64.8 & \textbf{62.2} & 0.85 & 1.90 & 35.1 & \textbf{53.6} & 1.40 & 2.72 & 72.5 \\
\midrule

\multirow{4}{*}{\textbf{SD~3.5 Large}}
& Unsteered        & 58.4 & 0.78 & 1.50 & 43.9 & 44.8 & 0.98 & 1.56 & 24.1 & 38.9 & 1.55 & 2.54 & 43.8 \\
& \method{}-Static & 68.3 & 0.70 & 1.57 & 53.4 & 52.9 & 0.97 & 1.78 & 29.5 & 46.7 & 1.40 & 2.52 & 52.7 \\
& \method{}-Feedback & 71.4 & 0.63 & 1.51 & 54.9 & 59.1 & 0.80 & 1.77 & 32.3 & 51.1 & 1.31 & 2.44 & 58.7 \\
& \method{}-Adaptive & \textbf{78.3} & 0.41 & 1.05 & 62.0 & \textbf{65.6} & 0.73 & 1.60 & 38.1 & \textbf{56.1} & 1.16 & 2.32 & 70.4 \\
\midrule

\multirow{5}{*}{\textbf{SDXL}}
& Unsteered        & 31.7 & 2.47 & 4.92 & 9.4 & 26.5 & 3.09 & 5.85 & 5.3 & 19.7 & 4.02 & 7.94 & 9.4 \\
& CountGen         & 50.3 & 1.90 & 4.60 & 44.1 & 41.7 & 2.04 & 4.68 & 29.2 & 29.4 & 2.70 & 5.45 & 55.7 \\
& \method{}-Static & 39.8 & 1.98 & 3.85 & 11.2 & 31.8 & 2.40 & 4.19 & 6.3 & 27.5 & 2.61 & 4.62 & 11.2 \\
& \method{}-Feedback & 46.6 & 1.67 & 3.59 & 15.0 & 36.4 & 2.24 & 4.93 & 8.8 & 27.2 & 2.71 & 4.96 & 15.8 \\
& \method{}-Adaptive & \textbf{54.0} & 1.50 & 3.47 & 19.4 & \textbf{45.5} & 1.93 & 4.52 & 11.6 & \textbf{34.7} & 2.34 & 3.90 & 21.2 \\
\midrule

\multirow{1}{*}{\textbf{SD~1.4}}
& Counting Guidance
                   & 28.6 & 2.19 & 3.62 & 19.9 & 19.4 & 2.80 & 4.27 & 11.1 & 10.8 & 3.69 & 4.85 & 19.6 \\
\bottomrule
\end{tabular}
}
\end{table*}

\textbf{Baselines.}
We compare \method{} against unsteered diffusion sampling and two representative baselines from prior work: \emph{CountGen}~\cite{binyamin2025makeitcount} and \emph{Counting Guidance}~\cite{kang2025counting}.
Both baselines rely on model-specific components and are therefore not backbone-agnostic. As a result, we evaluate each method on the backbone it was originally designed for: CountGen on SDXL and Counting Guidance on SD~1.4.
Although Counting Guidance is evaluated on a different backbone, we include it as a reference due to its comparable inference-time overhead to \method{}-Adaptive on SDXL, enabling a meaningful comparison of accuracy--runtime trade-offs.

\begin{figure}[b]
\centering
\begin{minipage}[c]{0.44\linewidth}
    % \vspace{0pt}
    \centering
    \includegraphics[width=\linewidth]{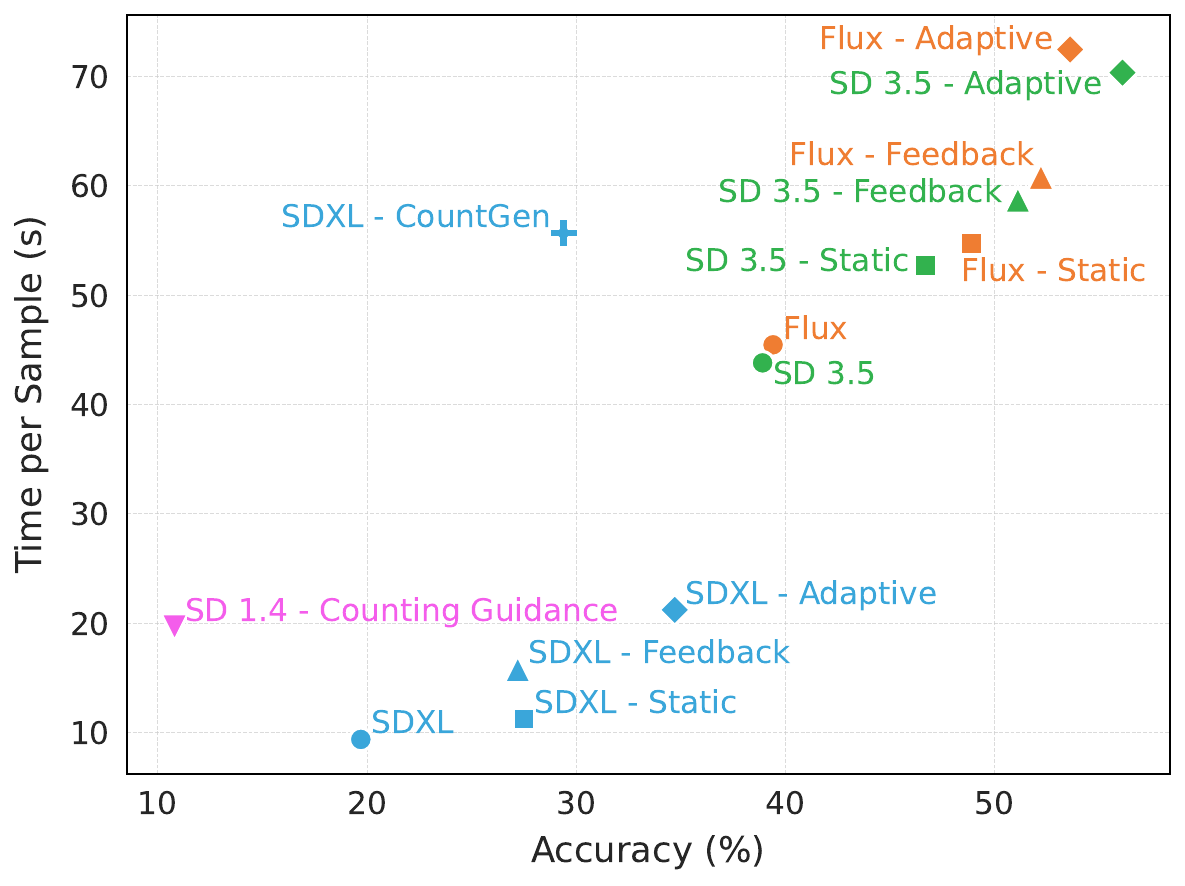}
    \caption{Accuracy–runtime trade-off on the \method{} dataset. Colors denote the diffusion backbone, while marker shapes indicate the method. \method{}-Adaptive achieves higher accuracy at comparable or lower runtime across diffusion backbones.}
    \label{fig:acc_time_athena}
\end{minipage}
\hfill
\begin{minipage}[c]{0.52\linewidth}
    % \vspace{0pt}
    \centering
    \includegraphics[width=\linewidth]{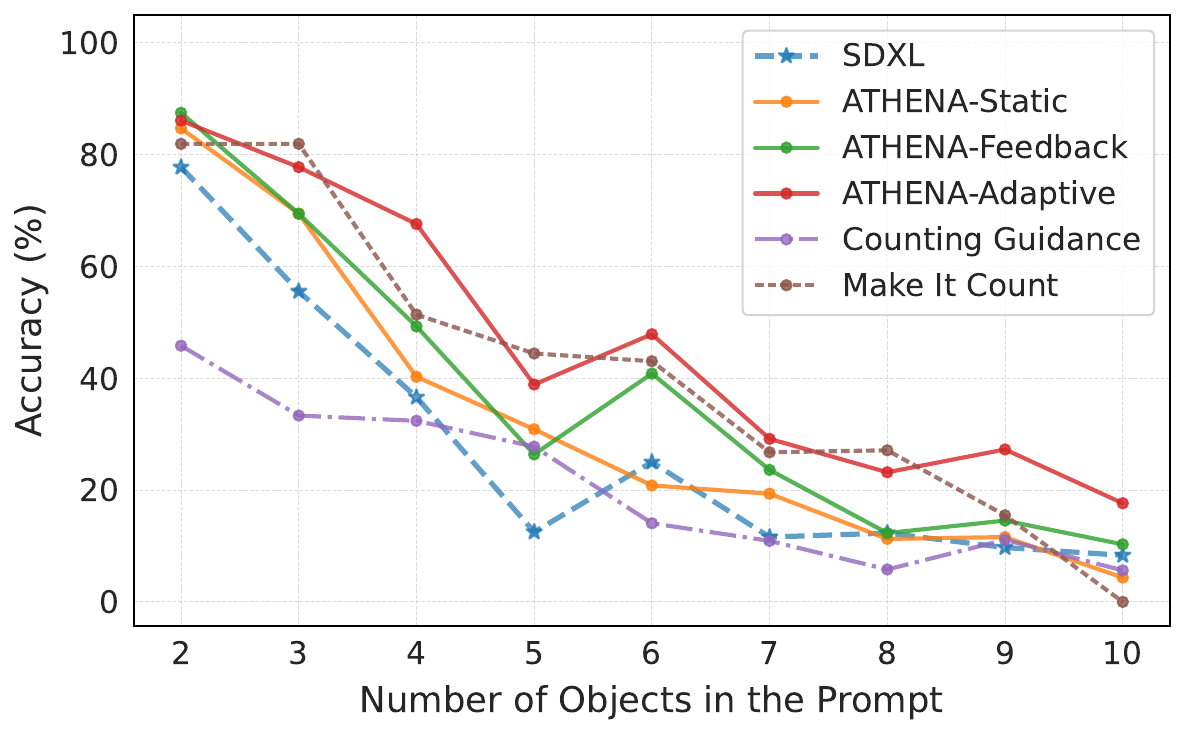}
    \caption{Counting accuracy versus target count on CoCoCount-E with SDXL backbone. \method{}-Adaptive achieves high accuracy for small counts and nearly $2\times$ the accuracy of unsteered sampling at larger counts.}
    \label{fig:acc_vs_count}
\end{minipage}
\end{figure}

% \begin{figure}[ht]
%     \centering
%     \includegraphics[width=0.65\linewidth]{figures/acc_time/acc_time_Athena.pdf}
%     \caption{Accuracy–runtime trade-off on the \method{} dataset. \method{}-Adaptive achieves higher accuracy at comparable or lower runtime across diffusion backbones.}
%     \label{fig:acc_time_athena}
% \end{figure}

CountGen does not support target counts above nine; for such prompts, we report unsteered sampling.
Counting Guidance does not support multi-word object categories; these are concatenated with underscores.

\textbf{Count Estimation.}
We estimate intermediate object counts using the pretrained open-vocabulary detector GroundingDINO~\cite{liu2024grounding}.
The detector is used only at test time for count estimation and is neither fine-tuned nor integrated into the generative models.
The same detector configuration is used across all methods and datasets.

\textbf{Hyperparameter Selection.}
All \method{} hyperparameters, including the estimation step $t_{\mathrm{est}}$, steering horizon $t_{\mathrm{steer}}$, and steering strength $\gamma$, are tuned exclusively on the CoCoCount dataset.
Hyperparameter tuning is conducted separately for each diffusion backbone.
Once selected, the resulting parameters are fixed and reused without modification for CoCoCount-E and the \method{} dataset.
Exact hyperparameter values are reported in \cref{appx:hyperparameters}.

% \begin{figure}[ht]
%     \centering
%     \includegraphics[width=0.82\linewidth]{figures/acc/acc_SDXL_CoCoCount_10.pdf}
%     \caption{Counting accuracy versus target count on CoCoCount-E with SDXL backbone. \method{}-Adaptive achieves high accuracy for small counts and nearly $2\times$ the accuracy of unsteered sampling at larger counts.}
%     \label{fig:acc_vs_count}
% \end{figure}

\textbf{Computational Setup.}
All experiments are conducted on a single GPU per run.
We use NVIDIA RTX A6000 GPUs (48\,GB) for CoCoCount and the \method{} dataset, and NVIDIA A100 SXM4 GPUs (40\,GB) for CoCoCount-E.
Within each dataset, the same GPU type and random seed are used across all models and methods to ensure fair comparisons.
Absolute generation times may differ across datasets due to hardware differences; accordingly, we focus on relative accuracy--runtime trade-offs with fixed hardware.

\begin{figure*}[!ht]
    \centering
    \includegraphics[width=0.99\linewidth]{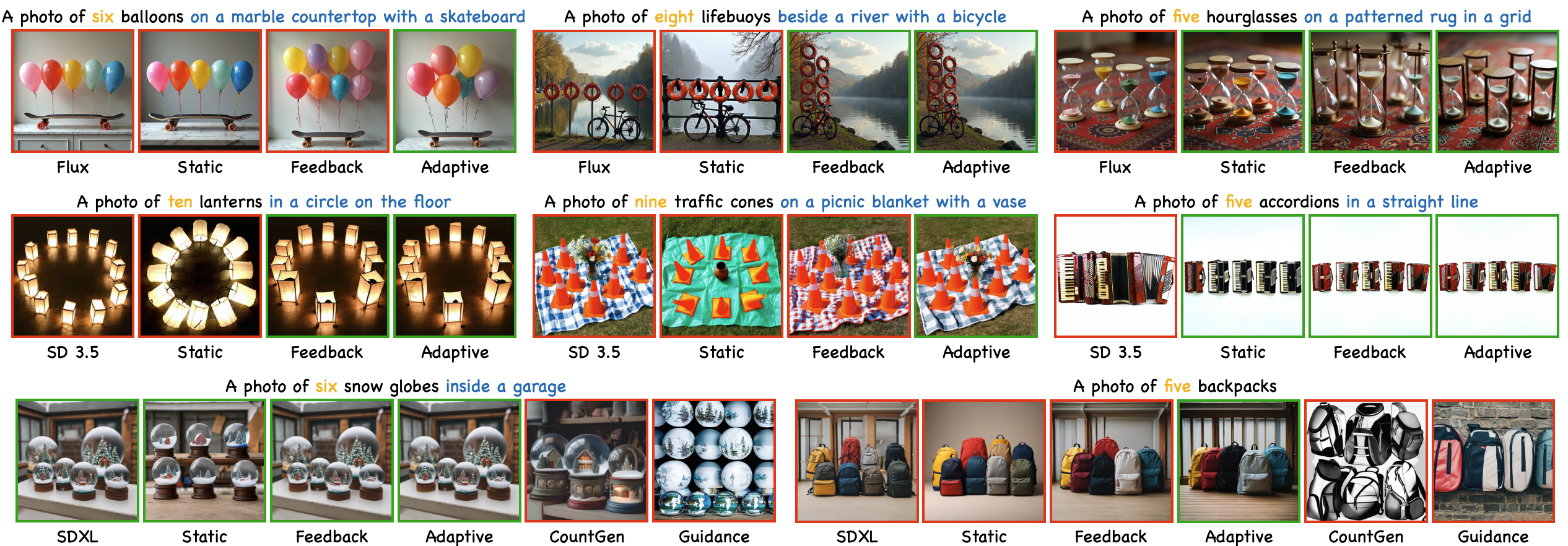}
    \caption{Qualitative results on complex, distractor-rich prompts. \emph{Guidance} denotes Counting Guidance. Green borders indicate correct counts; red borders indicate incorrect counts. All \method{} variants preserve scene structure and color consistency while handling relational and multi-object instructions. Compared to prior baselines, \method{} avoids visual artifacts and background distortion, with adaptive steering yielding the most accurate counts.
    }
    \label{fig:qual_results}
\end{figure*}

\subsection{Quantitative Results}
Table~\ref{tab:quant_results} summarizes counting performance and runtime across diffusion backbones and datasets.
\method{}-Adaptive consistently improves exact-count accuracy over unsteered generation, with absolute gains of at least 11.8\% (FLUX.1-dev on CoCoCount) and up to 22.3\% (SDXL on CoCoCount).
Adaptive steering outperforms static and feedback variants in nearly all cases; the only exception is CoCoCount with FLUX.1-dev, where static steering performs slightly better.
Overall, these results demonstrate that direction-aware adaptation of the steering strength is critical for robust count control across models and datasets.

Accuracy improvements are achieved with favorable efficiency trade-offs.
As shown in Table~\ref{tab:quant_results} and Figure~\ref{fig:acc_time_athena}, \method{}-Adaptive attains higher accuracy than CountGen while being over $2.5\times$ faster on the \method{} dataset.
At comparable runtime to Counting Guidance, \method{}-Adaptive on SDXL achieves nearly $2\times$ higher counting accuracy.
In terms of memory usage, CountGen requires 47.5\,GB, Counting Guidance uses 17.2\,GB, while \method{}-Adaptive on SDXL requires only 11.9\,GB.
Thus, \method{} achieves higher accuracy than prior baselines with approximately $4\times$ lower memory usage than CountGen, highlighting its efficiency as a lightweight, test-time control method.

We analyze counting accuracy as a function of the target object count.
Figure~\ref{fig:acc_vs_count} reports results on CoCoCount-E with the SDXL backbone.
Accuracy decreases as the target count increases for all methods, reflecting the growing difficulty of enforcing precise cardinality.
Unsteered generation degrades sharply beyond three objects, and prior baselines exhibit limited robustness as counts increase.

In contrast, \method{}-Adaptive maintains consistently higher accuracy across the full count range.
It achieves close to 80\% accuracy for two to three objects and nearly $2\times$ the accuracy of unsteered sampling for larger counts.
These results indicate that adaptive steering mitigates early semantic drift and slows error accumulation as counting difficulty increases.
Additional accuracy--count and accuracy--runtime figures are provided in \cref{appx:quantitative}.

\subsection{Qualitative Results}

Figure~\ref{fig:qual_results} presents comparisons between unsteered generation, baselines, and \method{} across diverse prompts.
As shown, \method{} improves count fidelity while preserving scene structure, object appearance, background coherence, and color consistency.
Unlike prior baselines, \method{} avoids visual artifacts such as background blurring, object blending, or disrupted spatial layout, and maintains coherent object–scene relationships in the shown examples, including cases with complex distractors and relational constraints.

In contrast, prior methods often modify scene content or introduce visual distortions when attempting to enforce counts.
CountGen frequently alters scene semantics, while Counting Guidance often fails to correct miscounting and degrades visual quality for complex prompts.
Among the variants, \method{}-Adaptive most consistently resolves counting errors, particularly for higher target counts and dense layouts.
Failure cases are rare and typically arise from inaccurate intermediate detection in especially challenging scenes.
Overall, these results indicate that \method{} enables accurate object counting at test time without sacrificing visual fidelity. Additional examples are provided in \cref{appx:qualitative}.
\section{Conclusion}

We introduced \method{}, a model-agnostic test-time steering framework for improving object count fidelity in T2I diffusion models without retraining or architectural changes. By estimating counts at an intermediate diffusion step and applying early prompt-based steering, \method{} corrects numerical errors before structural mistakes become fixed. Across three diffusion backbones, \method{}-Adaptive improves exact-count accuracy by up to 22.3\%, achieves close to 80\% accuracy for small target counts, and consistently outperforms prior baselines. These gains come with favorable efficiency: \method{} is nearly $2.5\times$ faster than CountGen, achieves almost twice the accuracy of Counting Guidance at similar runtime, and requires roughly $4\times$ less memory on SDXL. Overall, \method{} is an effective and efficient inference-time method for enforcing discrete numerical constraints in diffusion-based image generation while preserving visual quality.

% \newpage
% \section*{Acknowledgements}
% We would like to sincerely thank Professor Willie Neiswanger and Justin Cho for their insightful feedback and valuable guidance on this work. This research was also in part supported by AWS credits through an Amazon Faculty Research Award and a NAIRR Pilot Award. M. Soltanolkotabi and MS. Sepehri were supported by the USC–Capital One Center for Responsible AI and Decision Making in Finance (CREDIF) Fellowship. M. Soltanolkotabi is also supported by the Packard Fellowship in Science and Engineering, a Sloan Research Fellowship in Mathematics, an NSF-CAREER under award \#1846369, DARPA FastNICS program, and NSF-CIF awards \#1813877 and \#2008443. and NIH DP2LM014564-01. 

\section*{Acknowledgements}
We sincerely thank Willie Neiswanger and Justin Cho for their insightful feedback and valuable guidance. This work was partially supported by AWS credits through an Amazon Faculty Research Award, a NAIRR Pilot Award, and generous funding by Coefficient Giving. M. Soltanolkotabi and M. S. Sepehri were supported by the USC-Capital One Center for Responsible AI and Decision Making in Finance (CREDIF) Fellowship. M. Soltanolkotabi is also supported by the Packard Fellowship in Science and Engineering, a Sloan Research Fellowship in Mathematics, NSF CAREER Award \#1846369, DARPA FastNICS program, NSF CIF Awards \#1813877 and \#2008443, and NIH Award DP2LM014564-01.

\small
\bibliographystyle{ieeenat_fullname}
\bibliography{ref}
\normalsize

\newpage
\appendix
\onecolumn

\section{Algorithmic Details of \method{}}

\subsection{\method{}-Static}\label{appx:static}

\begin{figure*}[ht]
    \centering
    \includegraphics[width=0.8\linewidth]{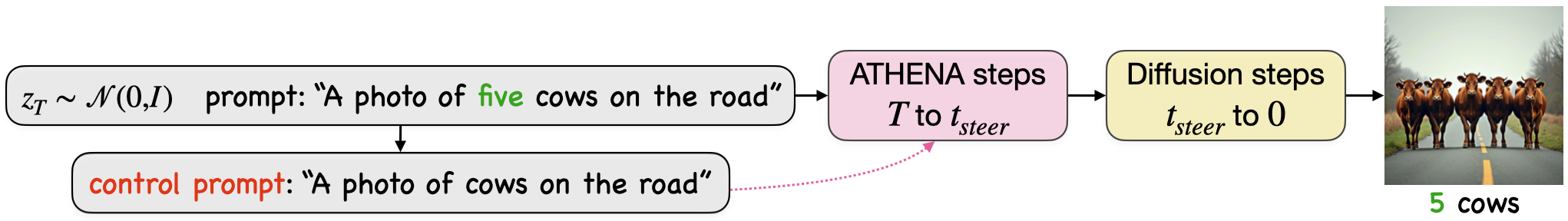}
    \caption{\method{}-Static pipeline. Starting from the same initial noise, prompt-based steering is applied using a fixed, count-agnostic control prompt from diffusion step $T$ down to a cutoff $t_{\mathrm{steer}}$. Standard diffusion then proceeds unmodified to completion, yielding improved count fidelity without intermediate feedback or adaptive adjustment.
    }
    \label{fig:athena_static}
\end{figure*}

% Modified by Berk. Also left some comments on my design choices:
% (Optional) maybe say final estimate is either the final image itself or it is passed through decoder if it is a latent diffusion model. It is not part of you algorithm really.
% In you figures you have "ATHENA" steps and "Diffusion" steps. It's better if it matches in the algorithm to track whats happening.
\begin{algorithm}[H]
\caption{\textsc{Diffusion-Steps}}
\label{alg:diffsion_step}
\small
\begin{algorithmic}[1]
\REQUIRE Denoiser: $\epsilon_\theta$, sampling operator: $\mathcal{S}_t$, text prompt: $p$, starting step: $t_s$, end step: $t_e$, initial latent: $z_{t_s}$

\FOR {$t = t_s$ to $t_e + 1$}
\STATE $\epsilon_t = \epsilon_{\theta}(t, z_t, p)$
\STATE $z_{t-1} = \mathcal{S}_t(z_t, \epsilon_{t})$
\ENDFOR
\STATE \textbf{Return:} $z_{t_e}$

\end{algorithmic}
\end{algorithm}

\begin{algorithm}[H]
\caption{\textsc{Athena-Steps}}
\label{alg:athena_step}
\small
\begin{algorithmic}[1]
\REQUIRE Denoiser: $\epsilon_\theta$, sampling operator: $\mathcal{S}_t$, text prompt: $p$, control prompt: $\hat{p}$, starting step: $t_s$, end step: $t_e$, initial latent: $z_{t_s}$, steering strength: $\gamma>0$

\FOR {$t = t_s$ to $t_e + 1$}
\STATE $\epsilon_t = \epsilon_{\theta}(t, z_t, p)$
\STATE $\hat{\epsilon}_t = \epsilon_{\theta}(t, z_t, \hat{p})$
\STATE $\epsilon_{steer} = \epsilon_t + \gamma( \epsilon_t - \hat{\epsilon}_t)$
\STATE $\epsilon_{steer} = \frac{\lVert \epsilon_t \rVert}{\lVert \epsilon_{steer} \rVert} \epsilon_{steer}$ \COMMENT{{Matching the norm of the new update with the original update}}
\STATE $z_{t-1} = \mathcal{S}_t(z_t, \epsilon_{steer})$
\ENDFOR
\STATE \textbf{Return:} $z_{t_e}$

\end{algorithmic}
\end{algorithm}

\begin{algorithm}[H]
\caption{\textsc{Athena-Static}}
\label{alg:athena_static}
\small
\begin{algorithmic}[1]
\REQUIRE Denoiser: $\epsilon_\theta$, sampling operator: $\mathcal{S}_t$, text prompt: $p$, target count: $k$, steering steps: $t_{steer}$, total steps: $T$, steering strength: $\gamma>0$, decoder $\mathrm{Dec}(\cdot)$
\STATE $z_T \sim \mathcal{N}(0, I)$
\STATE $\hat{p} \leftarrow \text{remove } k \text{ from } p$
\STATE $z_{t_{steer}} = \text{\textbf{ATHENA-Steps} ($\epsilon_\theta$, $\mathcal{S}_t$, $p$, $\hat{p}$, $t_s = T$, $t_e = t_{steer}$, $z_T$, $\gamma>0$)}$
\STATE $z_{0} = \text{\textbf{Diffusion-Steps} ($\epsilon_\theta$, $\mathcal{S}_t$, $p$, $t_s = t_{steer}$, $t_e = 0$, $z_{t_{steer}}$)}$
\STATE $x = \mathrm{Dec}(z_0)$
\STATE \textbf{Return:} $x$

\end{algorithmic}
\end{algorithm}

\newpage
\subsection{\method{}-Feedback}\label{appx:feedback}

\begin{figure*}[ht]
    \centering
    \includegraphics[width=0.38\linewidth]{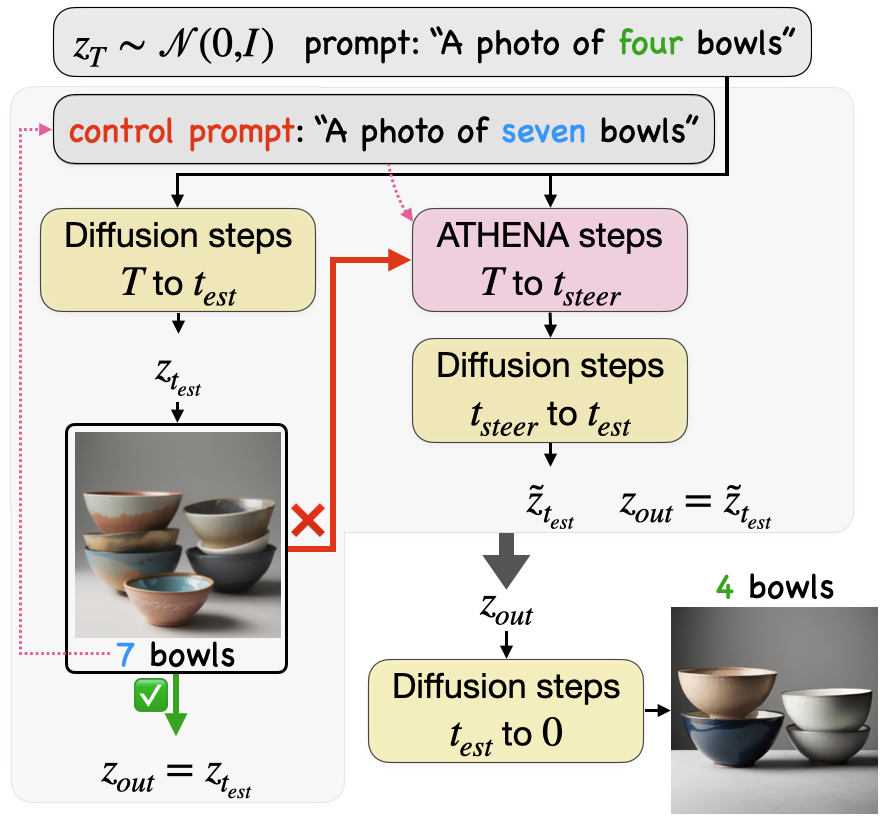}
    \caption{\method{}-Feedback pipeline.
    An intermediate count is estimated at step $t_{\mathrm{est}}$ without steering.
    If the estimate differs from the target, generation is restarted from the same initial noise using a feedback control prompt, and a single early-stage steering phase is applied before completing diffusion.}
    \label{fig:athena_feedback}
\end{figure*}

\begin{algorithm}[H]
\caption{\textsc{Athena-Feedback}}
\label{alg:athena_feedback}
\small
\begin{algorithmic}[1]
\REQUIRE Denoiser: $\epsilon_\theta$, sampling operator: $\mathcal{S}_t$, reconstruction $\mathcal{D}_t$, text prompt: $p$, target count: $k$, steering steps: $t_{steer}$, estimation step $t_{\mathrm{est}}$, total steps: $T$, steering strength: $\gamma>0$, decoder $\mathrm{Dec}(\cdot)$, counter $\mathrm{Count}(\cdot)$
\STATE $z_T \sim \mathcal{N}(0, I)$
\STATE $\hat{p} \leftarrow \text{remove } k \text{ from } p$
\STATE $z_{t_{est}} = \text{\textbf{Diffusion-Steps} ($\epsilon_\theta$, $\mathcal{S}_t$, $p$, $t_s = T$, $t_e = t_{est}$, $z_T$)}$
\STATE $\hat{z}_0=\mathcal{D}_t(z_{t_{est}})$
\STATE $\hat{x} = \mathrm{Dec}(\hat{z}_0)$
\STATE $c = \mathrm{Count}(\hat{x})$
\IF{$c = k$}
    \STATE $z_{out} = z_{t_{est}}$
\ELSE
    \STATE $\hat{p} \leftarrow \text{replace } k \text{ in } p \text{ with } c$
    \STATE $z_{t_{steer}} = \text{\textbf{ATHENA-Steps} ($\epsilon_\theta$, $\mathcal{S}_t$, $p$, $\hat{p}$, $t_s = T$, $t_e = t_{steer}$, $z_T$, $\gamma>0$)}$
    \STATE $\tilde{z}_{t_{est}} = \text{\textbf{Diffusion-Steps} ($\epsilon_\theta$, $\mathcal{S}_t$, $p$, $t_s = t_{steer}$, $t_e = t_{est}$, $z_{t_{est}}$)}$
    \STATE $z_{out} = \tilde{z}_{t_{est}}$
\ENDIF
\STATE $z_{0} = \text{\textbf{Diffusion-Steps} ($\epsilon_\theta$, $\mathcal{S}_t$, $p$, $t_s = t_{est}$, $t_e = 0$, $z_{out}$)}$
\STATE $x = \mathrm{Dec}(z_0)$
\STATE \textbf{Return:} $x$

\end{algorithmic}
\end{algorithm}

\newpage
\subsection{\method{}-Adaptive}\label{appx:adaptive}

\begin{figure}[ht]
    \centering
    \includegraphics[width=0.45\linewidth]{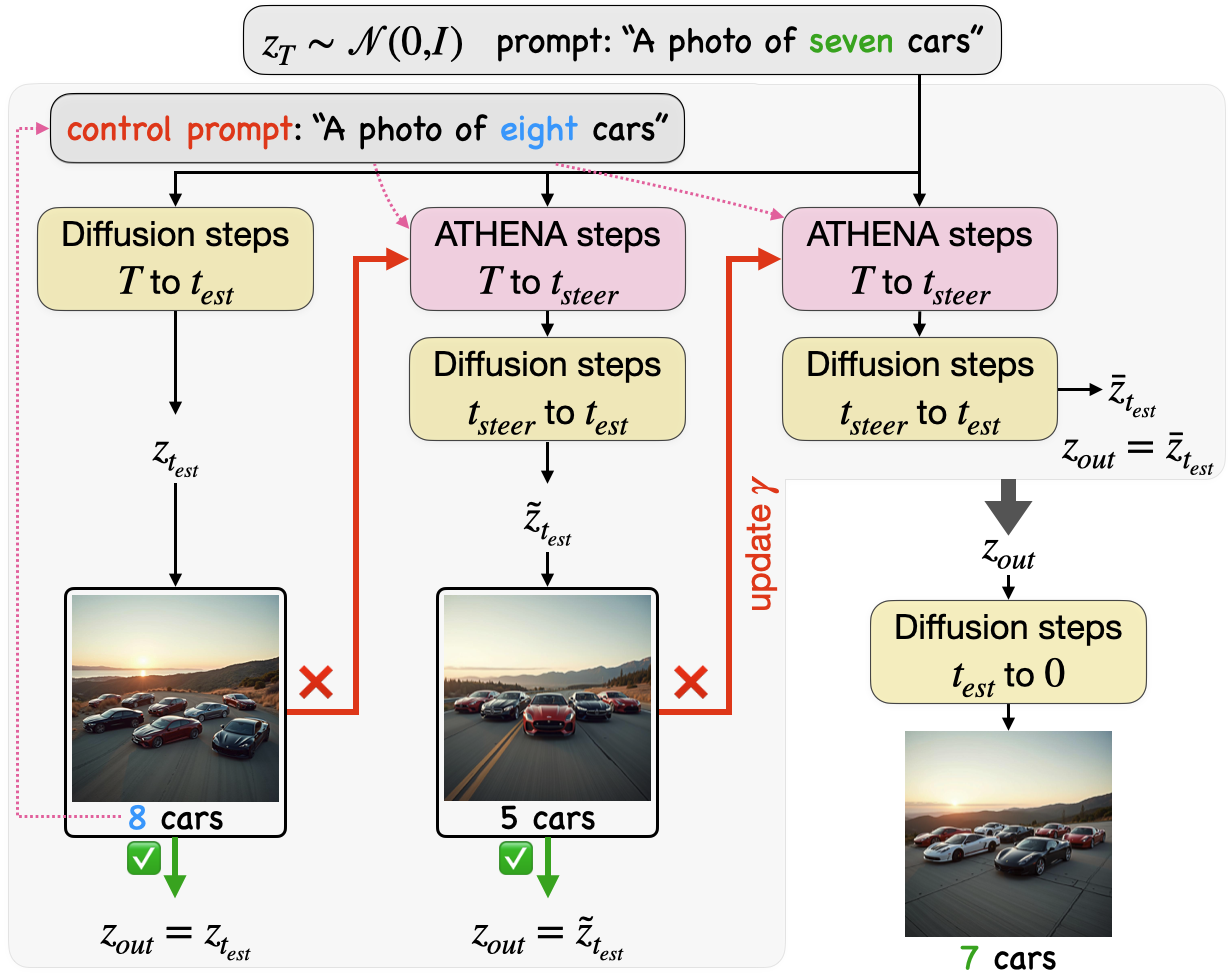}
    \caption{\method{}-Adaptive pipeline. The method estimates object count at an intermediate diffusion step, applies early-stage prompt-based steering, adaptively adjusts the steering strength once based on the observed error direction, and completes generation using standard diffusion.
    }
    \label{fig:athena_adaptive_apx}
\end{figure}

\begin{algorithm}[H]
\caption{\textsc{Athena-Adaptive}}
\label{alg:athena_adaptive}
\small
\begin{algorithmic}[1]
\REQUIRE Denoiser: $\epsilon_\theta$, sampling operator: $\mathcal{S}_t$, reconstruction $\mathcal{D}_t$, text prompt: $p$, target count: $k$, steering steps: $t_{steer}$, estimation step $t_{\mathrm{est}}$, total steps: $T$, steering strength: $\gamma>0$, decoder $\mathrm{Dec}(\cdot)$, counter $\mathrm{Count}(\cdot)$
\STATE $z_T \sim \mathcal{N}(0, I)$
\STATE $\hat{p} \leftarrow \text{remove } k \text{ from } p$
\STATE $z_{t_{est}} = \text{\textbf{Diffusion-Steps} ($\epsilon_\theta$, $\mathcal{S}_t$, $p$, $t_s = T$, $t_e = t_{est}$, $z_T$)}$
\STATE $\hat{z}_0=\mathcal{D}_t(z_{t_{est}})$
\STATE $\hat{x} = \mathrm{Dec}(\hat{z}_0)$
\STATE $c_1 = \mathrm{Count}(\hat{x})$
\IF{$c_1 = k$}
    \STATE $z_{out} = z_{t_{est}}$
\ELSE
    \STATE $\hat{p} \leftarrow \text{replace } k \text{ in } p \text{ with } c$
    \STATE $z_{t_{steer}} = \text{\textbf{ATHENA-Steps} ($\epsilon_\theta$, $\mathcal{S}_t$, $p$, $\hat{p}$, $t_s = T$, $t_e = t_{steer}$, $z_T$, $\gamma>0$)}$
    \STATE $\tilde{z}_{t_{est}} = \text{\textbf{Diffusion-Steps} ($\epsilon_\theta$, $\mathcal{S}_t$, $p$, $t_s = t_{steer}$, $t_e = t_{est}$, $z_{t_{est}}$)}$
    \STATE $\hat{z}_0=\mathcal{D}_t(\tilde{z}_{t_{est}})$
    \STATE $\hat{x} = \mathrm{Dec}(\hat{z}_0)$
    \STATE $c_2 = \mathrm{Count}(\hat{x})$
    \IF{$c_2 = k$}
        \STATE $z_{out} = \tilde{z}_{t_{est}}$
    \ELSE
        \IF{$(c_1 \leq c_2 < k) \lor (k < c_2 \leq c_1)$}
            \STATE \COMMENT{Case of not adding/removing enough objects}
            \STATE $\gamma \leftarrow 2\times\gamma$
        \ELSE
            \STATE \COMMENT{Case of adding/removing too many objects}
            \STATE $\gamma \leftarrow \frac{\gamma}{2}$
        \ENDIF
        \STATE $z_{t_{steer}} = \text{\textbf{ATHENA-Steps} ($\epsilon_\theta$, $\mathcal{S}_t$, $p$, $\hat{p}$, $t_s = T$, $t_e = t_{steer}$, $z_T$, $\gamma>0$)}$
        \STATE $\bar{z}_{t_{est}} = \text{\textbf{Diffusion-Steps} ($\epsilon_\theta$, $\mathcal{S}_t$, $p$, $t_s = t_{steer}$, $t_e = t_{est}$, $z_{t_{est}}$)}$
        \STATE $z_{out} = \bar{z}_{t_{est}}$
    \ENDIF
\ENDIF
\STATE $z_{0} = \text{\textbf{Diffusion-Steps} ($\epsilon_\theta$, $\mathcal{S}_t$, $p$, $t_s = t_{est}$, $t_e = 0$, $z_{out}$)}$
\STATE $x = \mathrm{Dec}(z_0)$
\STATE \textbf{Return:} $x$

\end{algorithmic}
\end{algorithm}

\newpage
\section{Dataset Construction and Statistics}
\label{appx:datasets}

\subsection{CoCoCount and CoCoCount-E}
\label{appx:cococount_details}

The CoCoCount dataset~\cite{binyamin2025makeitcount} consists of text prompts with explicit numerical constraints, targeting object counts from 2 to 10.
The original release includes prompts that differ only by random generation seed, as well as prompts with identical semantic structure but different target counts.
As a result, the dataset contains repeated prompt content and heterogeneous count distributions.

For evaluation, we construct a deduplicated version of CoCoCount by retaining a single instance of each unique prompt text.
This results in 161 distinct prompts, which we use as the CoCoCount benchmark in all experiments.

To enable controlled analysis across target cardinalities, we further construct \emph{CoCoCount-E}, an extended benchmark derived from CoCoCount.
Starting from a filtered subset of 72 unique prompt templates, we systematically instantiate each prompt with target counts from 2 to 10 while keeping all other prompt content fixed.
This yields 648 prompts with consistent structure and a balanced count distribution.

Following prior work~\cite{binyamin2025makeitcount}, we restrict the target count range to 2--10, as generation quality and counting accuracy degrade substantially beyond this range.
We observe the same trend across all evaluated diffusion backbones.

\begin{table*}[h]
\caption{Dataset statistics for CoCoCount and CoCoCount-E.}
\centering
\small
\begin{tabular}{lccc}
\toprule
\textbf{Dataset} & \textbf{Unique Prompts} & \textbf{Count Range} & \textbf{Total Prompts} \\
\midrule
CoCoCount (deduplicated) & 161 & 2--10 & 161 \\
CoCoCount-E & 72 & 2--10 & 648 \\
\bottomrule
\end{tabular}
\end{table*}

\newpage
\subsection{\method{} Dataset Generation}
\label{appx:athena}

The \method{} dataset is generated by querying a large language model (GPT-5.2) using the prompt shown below.
This prompt instructs the model to produce object-counting instructions with controlled variations in object category, scene context, distractors, and relational constraints, while sampling target counts between 2 and 10.
Applying this prompt yields a structured collection of prompts organized into four difficulty levels, with consistent formatting and systematic increases in complexity. Finally, all generated samples are manually inspected to ensure quality.

\begin{promptbox}
\small
\paragraph{\large{Task Definition}}\mbox{}
\vspace{5pt}
\begin{Verbatim}[breaklines=true, breakanywhere=true]
Generate a challenging object-counting evaluation dataset for text-to-image diffusion models. The dataset is intended to evaluate count fidelity only, under increasingly difficult prompt conditions.

You are given:
- A JSON file containing the original CoCoCount dataset (attached).
- The specifications below.

You must generate a new dataset that:
- Does NOT reuse any object categories appearing in CoCoCount for Level 1.
- Uses clear, unambiguous noun-phrase prompts suitable for automatic counting.
- Contains no duplicate prompts.
- Contains exactly one counted object category per prompt.
- All prompts must begin with: "A photo of …"
\end{Verbatim}
\mbox{}\paragraph{\large{Dataset Output Format}}\mbox{}
\begin{Verbatim}[breaklines=true, breakanywhere=true]
- Output a single JSON file as a list of entries.
- Each entry must contain the following required fields.

Required JSON template (exact keys required):
\end{Verbatim}
\begin{CodeFence}
\begin{Verbatim}[breaklines=true, breakanywhere=true]
{
  "id": <unique integer>,
  "prompt": <string>,                 // full prompt, starts with "A photo of"
  "object": <string>,                 // singular object name
  "object_plural": <string>,          // plural form used in the prompt
  "number": <string>,                 // number in words (e.g., "three")
  "int_number": <integer>,                 // numeric count (e.g., 3)

  "level": <integer>,                 // 1, 2, 3, or 4
  "difficulty_tag": <string>,         // short descriptor (e.g., "hard-object", "scene", "distractor", "relation")

  "scene": <string | null>,           // scene phrase if present (e.g., "on the ground")
  "distractor": <string | null>,      // uncounted object if present
  "relation": <string | null>,        // relational phrase if present

  "source": "ATHENA"
}
\end{Verbatim}
\end{CodeFence}
\mbox{}\paragraph{\large{Detector Compatibility Constraint}}\mbox{}
\begin{Verbatim}[breaklines=true, breakanywhere=true]

All object categories MUST be reliably detectable by open-vocabulary grounding models
(e.g., Grounding DINO) under standard confidence thresholds.

Objects must:
- be visually salient at typical image resolutions,
- have strong and common visual-text associations,
- be distinguishable without fine-grained or microscopic detail,
commonly appear in natural images rather than technical diagrams or product-only photos.

Avoid objects that are:
- extremely small or thin,
- rare, niche, or domain-specific,
- visually indistinguishable without context,
- typically embedded inside other objects.
\end{Verbatim}
\mbox{}\paragraph{\large{Count Constraints}}\mbox{}
\begin{Verbatim}[breaklines=true, breakanywhere=true]
- Valid counts: 2 - 10
- Counts should be approximately balanced
- "number" must be the word form of "count"
- "object_plural" must match the prompt text exactly
\end{Verbatim}
\mbox{}\paragraph{\large{Difficulty Levels}}\mbox{}
\paragraph{Level 1: Hard Object Categories}\mbox{}
\begin{Verbatim}[breaklines=true, breakanywhere=true]
- Object categories must NOT appear in CoCoCount.
- Objects should be visually challenging for counting (e.g., instance ambiguity, moderate occlusion, reflective or deformable surfaces), but must remain clearly recognizable as distinct object instances by open-vocabulary detection models.
- No verbs
- No scene
- No distractors
\end{Verbatim}
\textbf{Prompt format:}
\begin{CodeFence}
\begin{Verbatim}[breaklines=true, breakanywhere=true]
A photo of <number> <object_plural>
\end{Verbatim}
\end{CodeFence}
\textbf{Example pattern:}
\emph{A photo of seven kites} \\
% \begin{CodeFence}
% \begin{Verbatim}[breaklines=true, breakanywhere=true]
% A photo of <number> <object_plural>
% \end{Verbatim}
% \end{CodeFence}
% \begin{Verbatim}[breaklines=true, breakanywhere=true]
% Example pattern:
% A photo of seven kites
% \end{Verbatim}
\mbox{}\paragraph{Level 2: Hard Objects + Scene Context}\mbox{}
\begin{Verbatim}[breaklines=true, breakanywhere=true]
- Same object constraints as Level 1.
- Add a scene phrase.
- Scene must not introduce additional countable objects.
- No verbs
- No distractors
\end{Verbatim}
\textbf{Prompt format:}
\begin{CodeFence}
\begin{Verbatim}[breaklines=true, breakanywhere=true]
A photo of <number> <object_plural> <scene>
\end{Verbatim}
\end{CodeFence}
\textbf{Example pattern:}
\emph{A photo of six lanterns in a temple hall} \\
% \begin{CodeFence}
% \begin{Verbatim}[breaklines=true, breakanywhere=true]
% A photo of <number> <object_plural> <scene>
% \end{Verbatim}
% \end{CodeFence}
% \begin{Verbatim}[breaklines=true, breakanywhere=true]
% Example pattern:
% A photo of six lanterns in a temple hall
% \end{Verbatim}
\mbox{}\paragraph{Level 3: Counted Object + Semantic Distractor}\mbox{}
\begin{Verbatim}[breaklines=true, breakanywhere=true]
- One counted object category.
- One uncounted distractor (no number specified).
- No verbs
- Avoid ambiguity about which object is counted.
\end{Verbatim}
\textbf{Prompt format:}
\begin{CodeFence}
\begin{Verbatim}[breaklines=true, breakanywhere=true]
A photo of <number> <object_plural> <scene> with <distractor>
\end{Verbatim}
\end{CodeFence}
\textbf{Example pattern:}
\emph{A photo of five apples on a table with a dog} \\
% \begin{Verbatim}[breaklines=true, breakanywhere=true]
% Example pattern:
% A photo of five apples on a table with a dog
% \end{Verbatim}
\mbox{}\paragraph{Level 4: Relational Language}\mbox{}
\begin{Verbatim}[breaklines=true, breakanywhere=true]
- One counted object category.
- Introduce spatial or relational structure.
- No second counted object.
- Relational phrasing may include light verb-based constructions if unavoidable.
\end{Verbatim}
\textbf{Prompt format:}
\begin{CodeFence}
\begin{Verbatim}[breaklines=true, breakanywhere=true]
A photo of <number> <object_plural> <relation>
\end{Verbatim}
\end{CodeFence}
\textbf{Example pattern:}
\emph{A photo of eight chairs arranged around a round table} \\
% \begin{CodeFence}
% \begin{Verbatim}[breaklines=true, breakanywhere=true]
% A photo of <number> <object_plural> <relation>
% \end{Verbatim}
% \end{CodeFence}

% \begin{Verbatim}[breaklines=true, breakanywhere=true]
% Example pattern:
% A photo of eight chairs arranged around a round table
% \end{Verbatim}
\mbox{}\paragraph{\large{Strict Constraints}}\mbox{}
\begin{Verbatim}[breaklines=true, breakanywhere=true]
- Do NOT reuse Level-1 object categories from CoCoCount.
- Do NOT include multiple counted objects.
- Do NOT use vague quantifiers ("several", "many").
- Do NOT use negation-based counting.
- Do NOT generate prompts requiring subjective interpretation.
- Do NOT generate near-duplicate prompts.
- Ensure "scene", "distractor", and "relation" are null when not applicable.
\end{Verbatim}
\mbox{}\paragraph{\large{Dataset Size}}\mbox{}
\begin{Verbatim}[breaklines=true, breakanywhere=true]
Generate exactly 360 prompts, distributed as:
- Level 1: 120
- Level 2: 100
- Level 3: 100
- Level 4: 40
\end{Verbatim}
\mbox{}\paragraph{\large{Output Requirements}}\mbox{}
\begin{Verbatim}[breaklines=true, breakanywhere=true]
- Output valid JSON only
- All required fields must be present
- Prompts must be natural, fluent, and concise
- Share the dataset link when you finish generating it
\end{Verbatim}
\end{promptbox}

\section{Hyperparameter Settings}
\label{appx:hyperparameters}

All diffusion models are evaluated using a fixed random seed (23) for image generation to ensure reproducibility.
Table~\ref{tab:hyperparameters} summarizes the hyperparameter settings used for \method{} across diffusion backbones and steering variants.

\begin{table*}[h]
\centering
\caption{Hyperparameter settings used for \method{} across diffusion backbones and steering variants.}
\label{tab:hyperparameters}
% \resizebox{0.8\linewidth}{!}{
\small
\begin{tabular}{lccc ccc ccc}
\toprule
& \multicolumn{9}{c}{\textbf{Diffusion Backbone}} \\
\cmidrule(lr){2-10}
\textbf{Parameter} 
& \multicolumn{3}{c}{\textbf{FLUX.1-dev}} 
& \multicolumn{3}{c}{\textbf{SD 3.5 Large}} 
& \multicolumn{3}{c}{\textbf{SDXL}} \\
\cmidrule(lr){2-4} \cmidrule(lr){5-7} \cmidrule(lr){8-10}
& Static & Feedback & Adaptive
& Static & Feedback & Adaptive
& Static & Feedback & Adaptive \\
\midrule
$t_{\mathrm{est}}$   & -- & 20 & 20 & -- & 20 & 20 & -- & 30 &  30 \\
$t_{\mathrm{steer}}$ & 10 & 5 & 5 & 10 & 5 & 5 & 10 & 10 & 10 \\
$\gamma$             & 4 & 4 & 4 & 4 & 4 & 4 & 5 & 5 & 5 \\
\bottomrule
\end{tabular}
% }
\end{table*}

\newpage
\section{Additional Experiments}

\subsection{Quantitative Results}\label{appx:quantitative}

\begin{figure*}[h]
    \centering
    \begin{subfigure}[t]{0.42\linewidth}
        \centering
        \includegraphics[width=\linewidth]{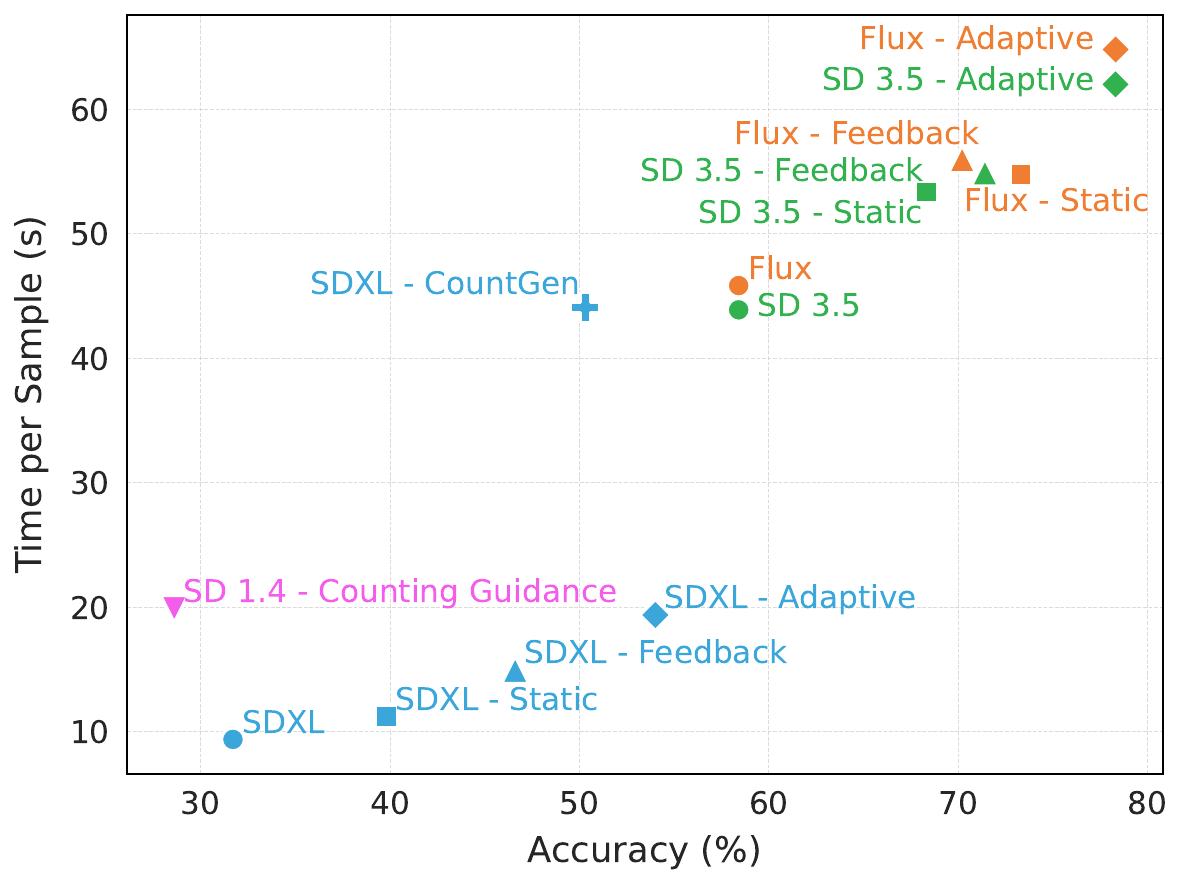}
        \caption{CoCoCount}
        \label{fig:acc_time_cococount}
    \end{subfigure}
    % \hfill
    \hspace{3em}
    \begin{subfigure}[t]{0.42\linewidth}
        \centering
        \includegraphics[width=\linewidth]{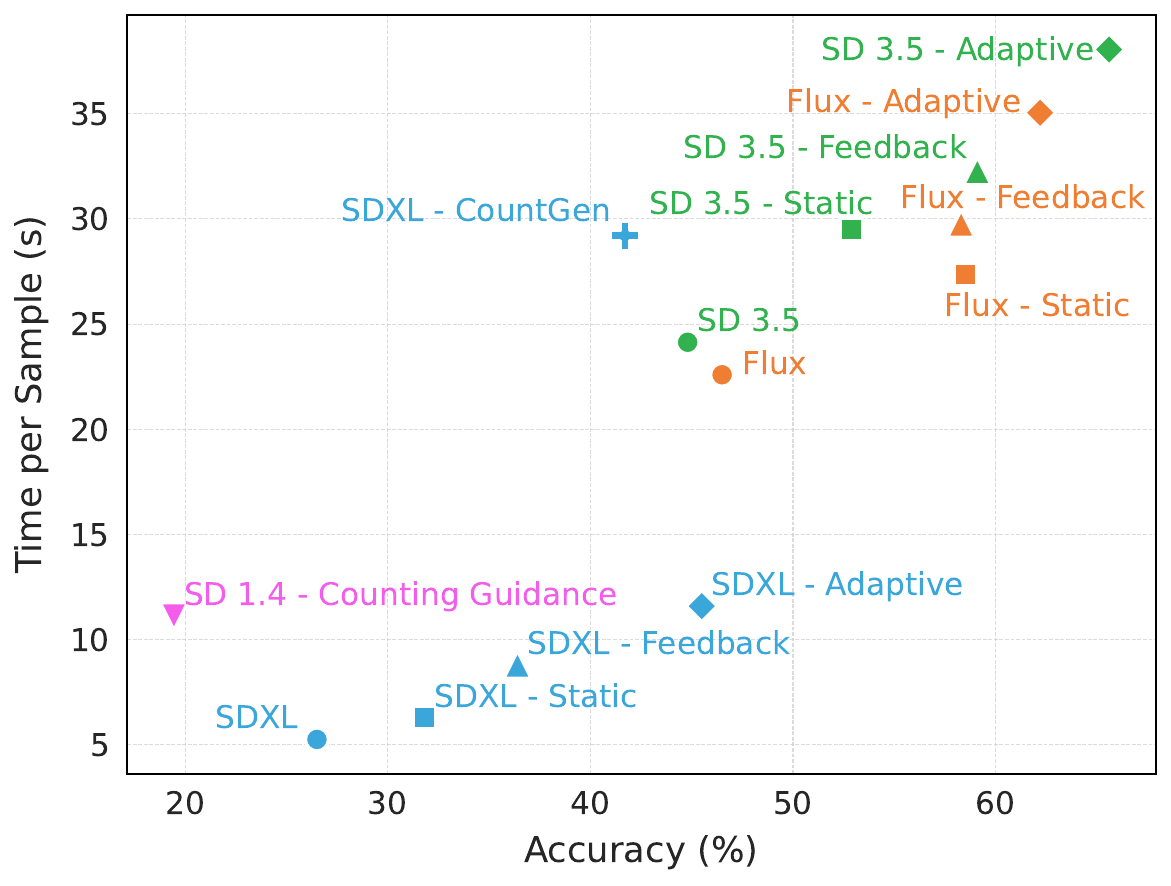}
        \caption{CoCoCount-E}
        \label{fig:acc_time_coccocountE}
    \end{subfigure}
    \caption{Accuracy--runtime trade-offs on CoCoCount and CoCoCount-E.  Colors denote the diffusion backbone, while marker shapes indicate the method. \method{}-Adaptive consistently achieves higher accuracy at comparable or lower runtime than prior baselines.}
    \label{fig:acc_time_appx}
\end{figure*}

\begin{figure*}[h]
    \centering
    \begin{subfigure}[t]{0.32\linewidth}
        \centering
        \includegraphics[width=\linewidth]{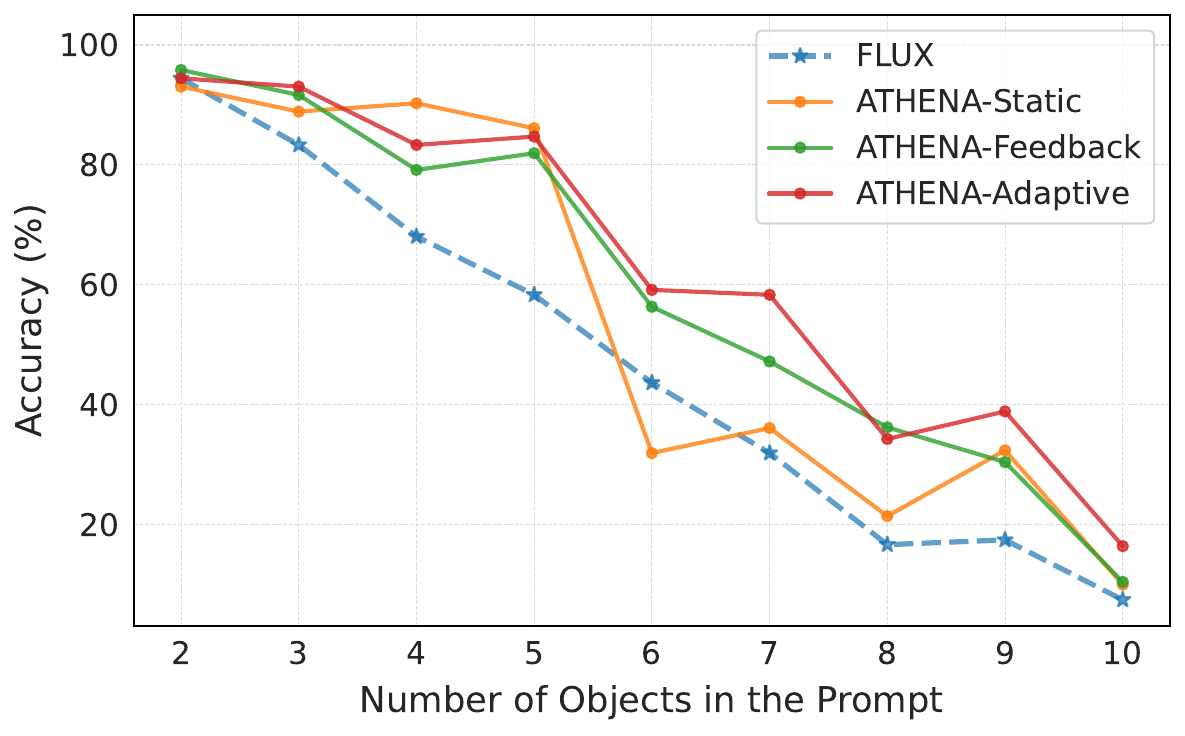}
        \caption{CoCoCount-E (FLUX)}
    \end{subfigure}
    \hfill
    \begin{subfigure}[t]{0.32\linewidth}
        \centering
        \includegraphics[width=\linewidth]{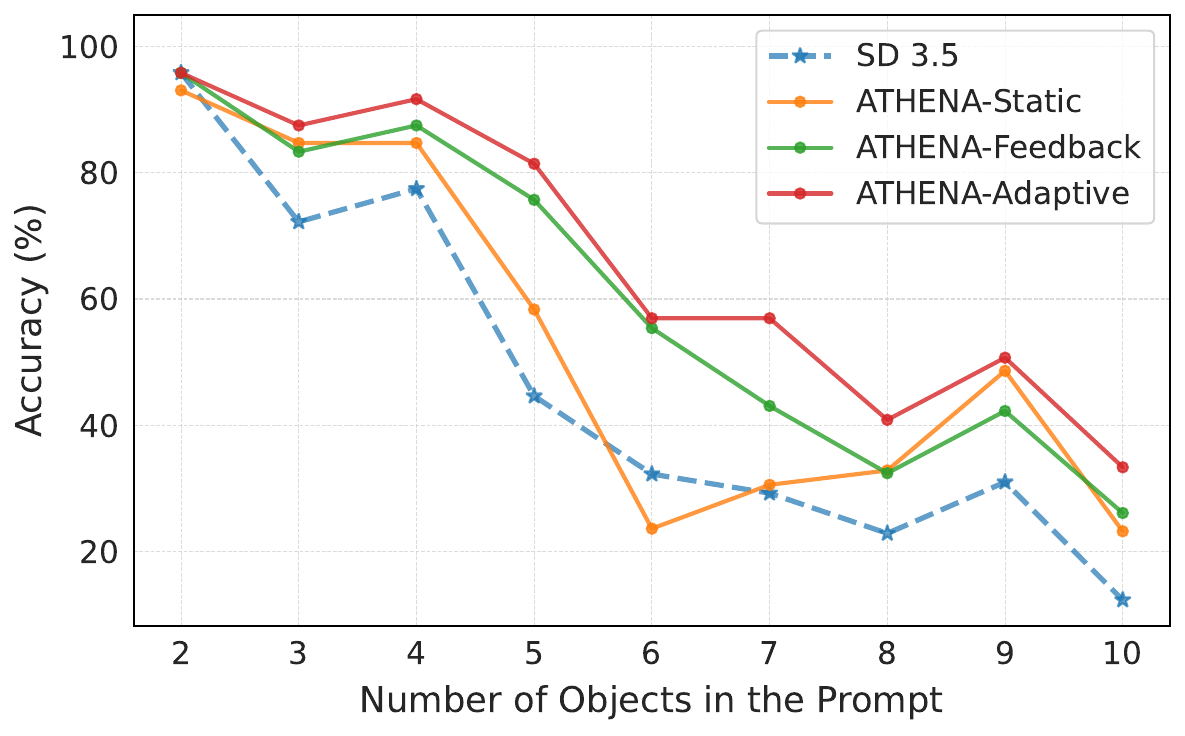}
        \caption{CoCoCount-E (SD 3.5)}
    \end{subfigure}
    \hfill
    \begin{subfigure}[t]{0.32\linewidth}
        \centering
        \includegraphics[width=\linewidth]{figures/acc/acc_SDXL_CoCoCount_10.pdf}
        \caption{CoCoCount-E (SDXL)}
    \end{subfigure}
    \begin{subfigure}[t]{0.32\linewidth}
        \centering
        \includegraphics[width=\linewidth]{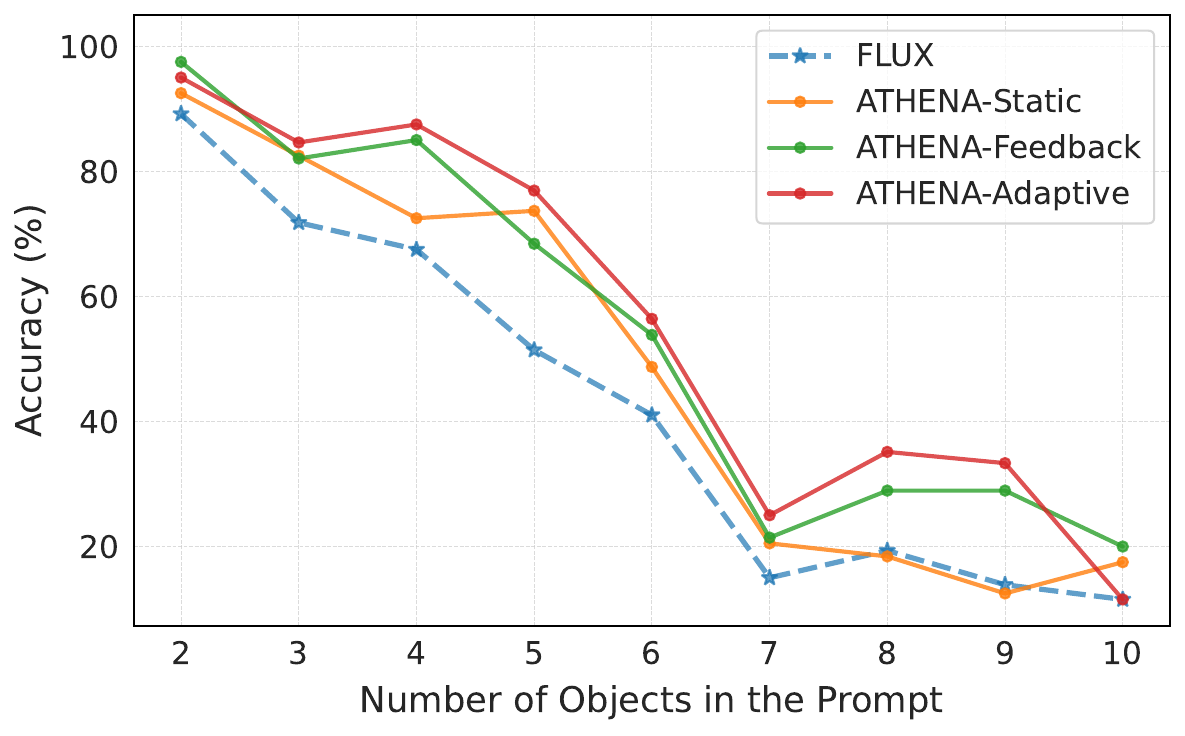}
        \caption{\method{} dataset (FLUX)}
    \end{subfigure}
    \hfill
    \begin{subfigure}[t]{0.32\linewidth}
        \centering
        \includegraphics[width=\linewidth]{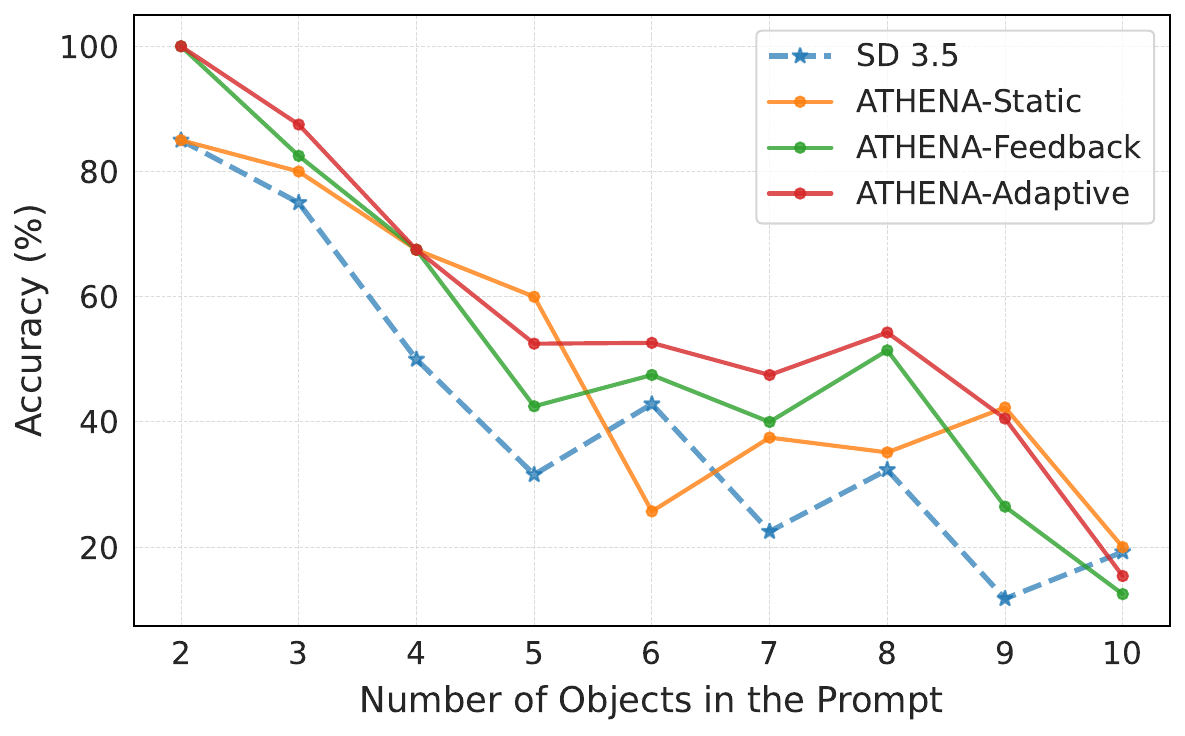}
        \caption{\method{} dataset (SD 3.5)}
    \end{subfigure}
    \hfill
    \begin{subfigure}[t]{0.32\linewidth}
        \centering
        \includegraphics[width=\linewidth]{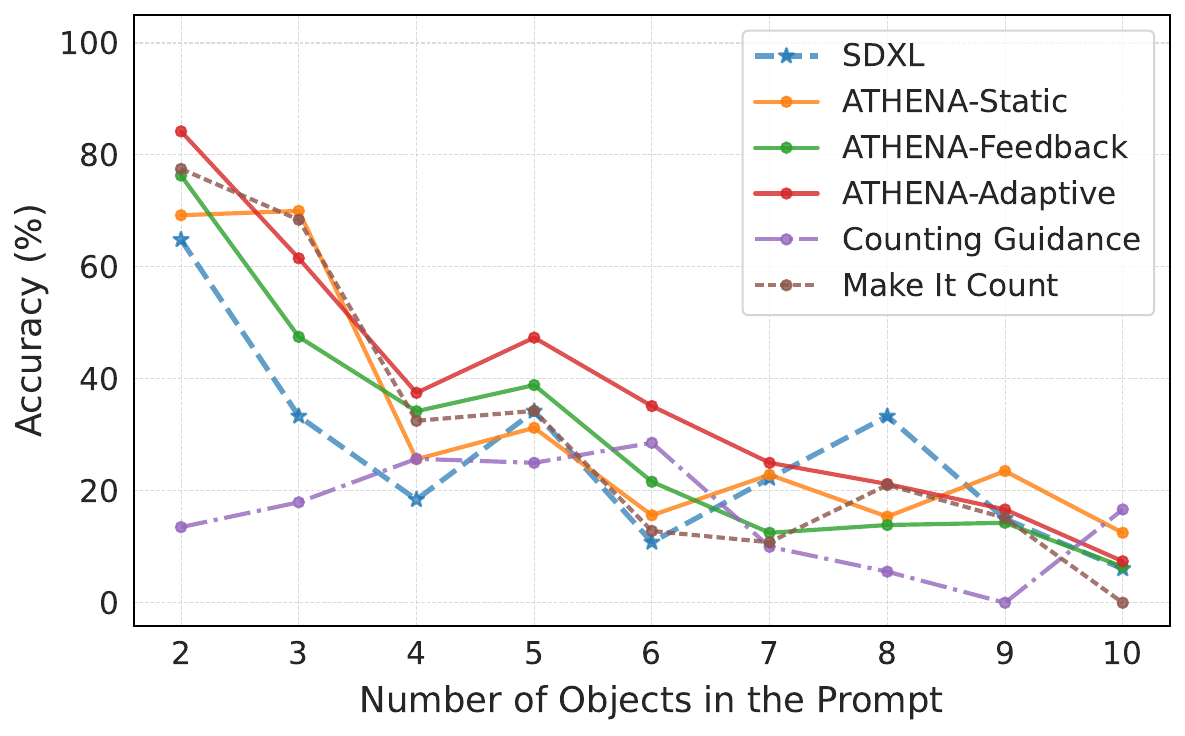}
        \caption{\method{} dataset (SDXL)}
    \end{subfigure}
    \caption{Counting accuracy versus target object count across datasets and diffusion backbones. \method{}-Adaptive consistently maintains higher accuracy as the target count increases, demonstrating improved robustness to increasing counting difficulty.}
    \label{fig:acc_count_appx}
\end{figure*}

\newpage
\subsection{Qualitative Results}\label{appx:qualitative}
\begin{figure*}[h]
    \centering
    \includegraphics[width=0.99\linewidth]{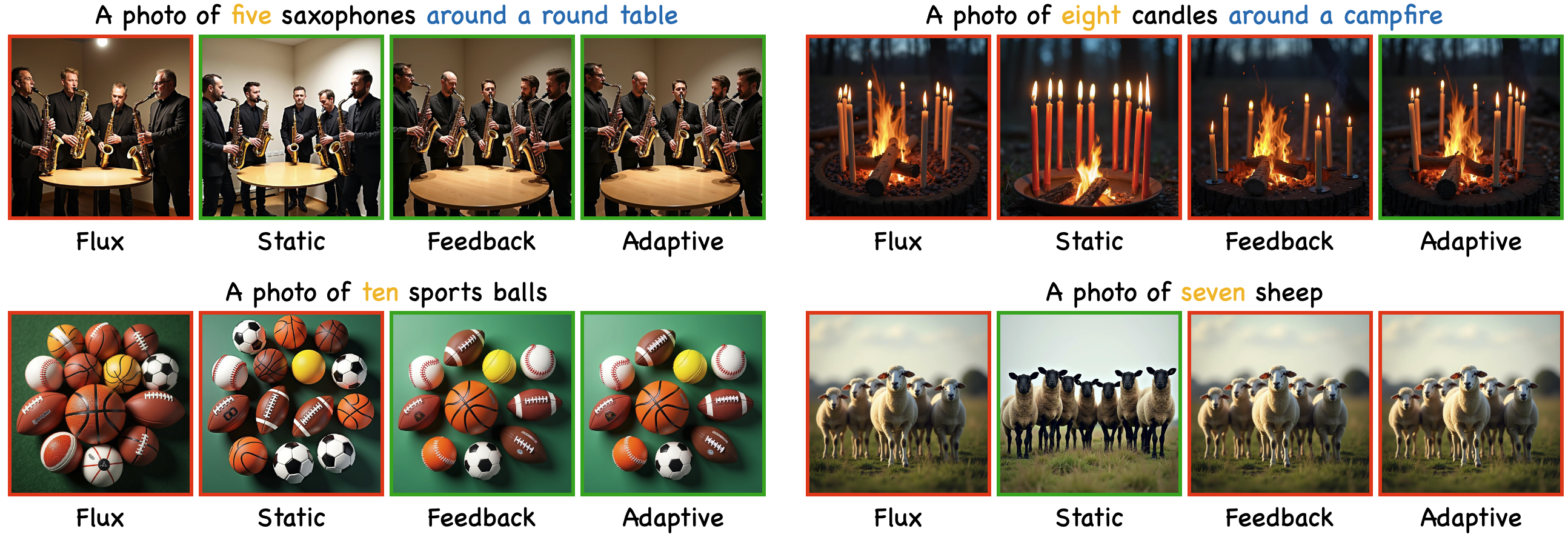}
    \caption{Additional qualitative results on relational and multi-object prompts. Results are shown for unsteered generation (FLUX) and the three \method{} variants. Green borders indicate correct object counts, while red borders denote counting errors. \method{} improves count fidelity across diverse settings (e.g., grouping, circular layouts, and natural scenes) while preserving scene structure and visual coherence, with the adaptive variant yielding the most consistent corrections.
    }
    \label{fig:examples_flux}
\end{figure*}

\begin{figure*}[h]
    \centering
    \includegraphics[width=0.99\linewidth]{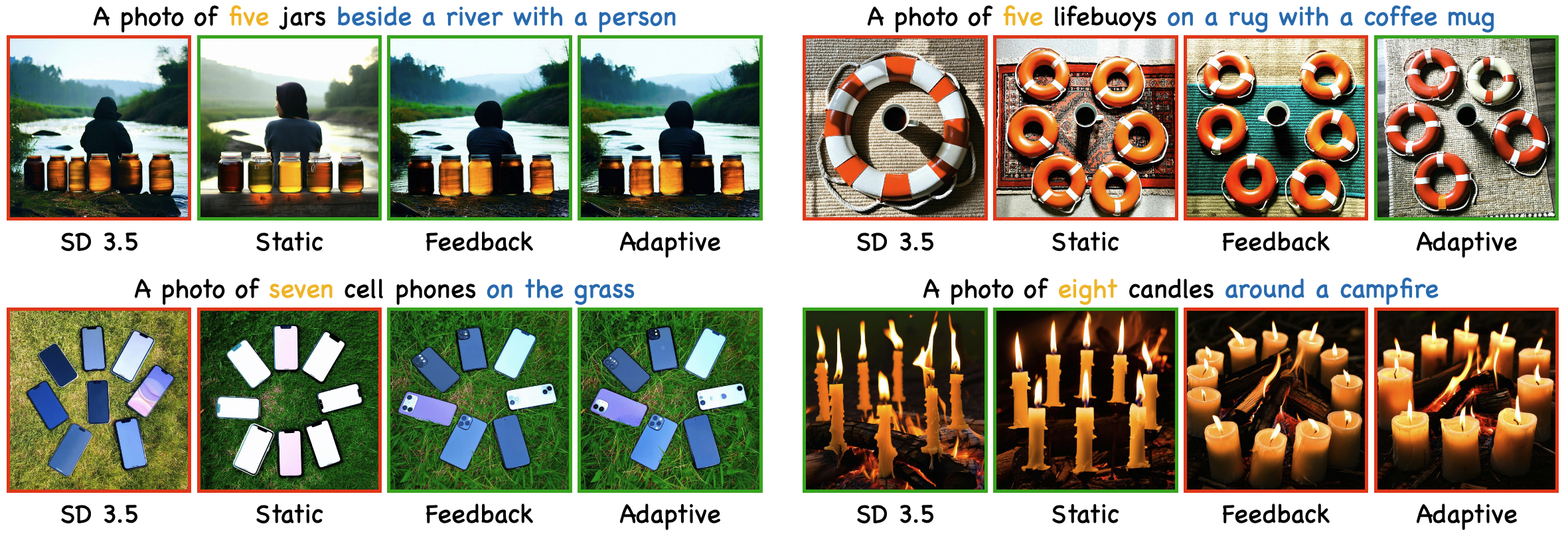}
    \caption{Additional qualitative results on relational and multi-object prompts. Results are shown for unsteered generation (SD 3.5) and the three \method{} variants. Green borders indicate correct object counts, while red borders denote counting errors. \method{} improves count fidelity across diverse settings (e.g., grouping, circular layouts, and natural scenes) while preserving scene structure and visual coherence, with the adaptive variant yielding the most consistent corrections.
    }
    \label{fig:examples_sd}
\end{figure*}

\begin{figure*}[ht]
    \centering
    \includegraphics[width=0.75\linewidth]{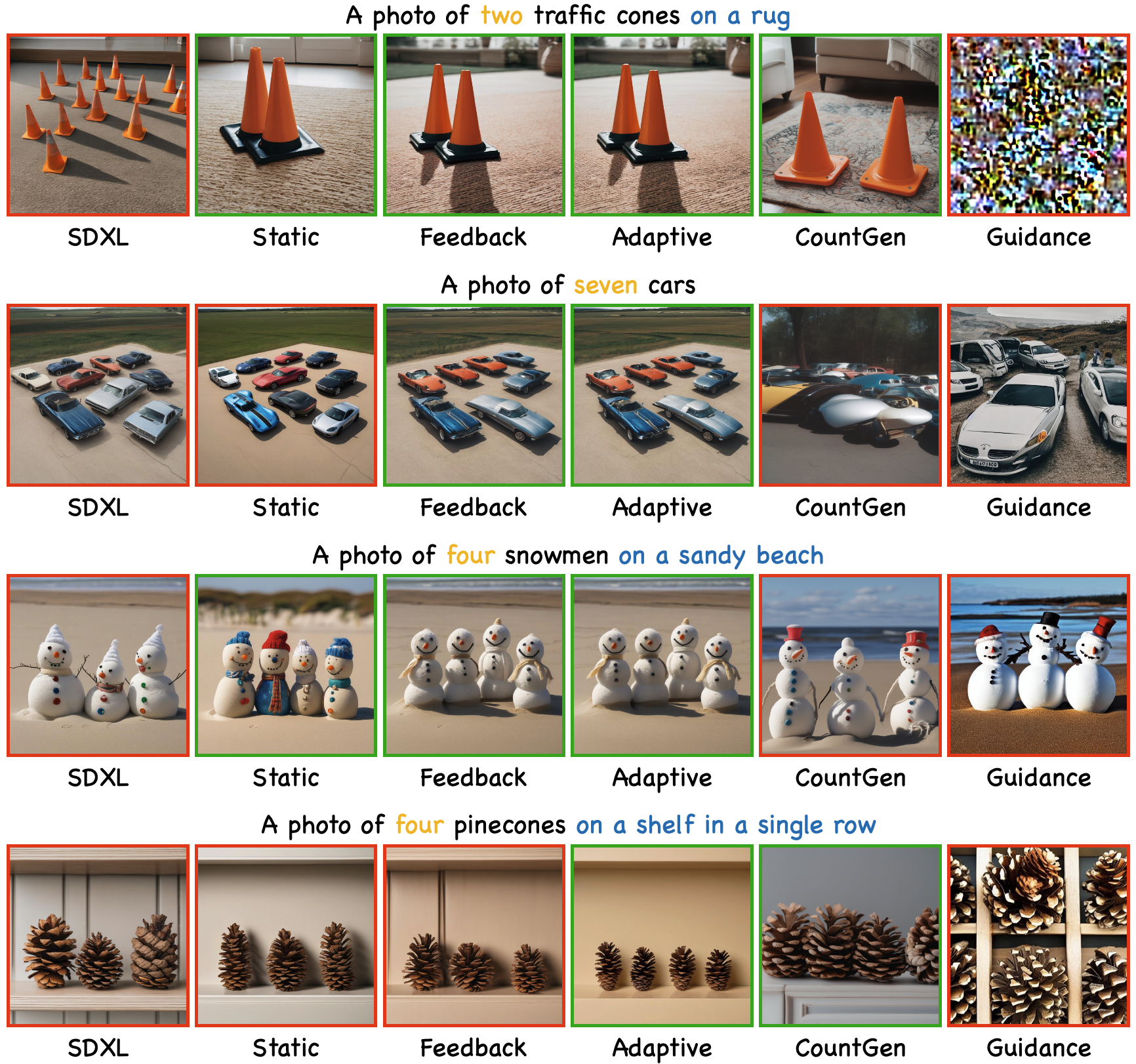}
        \caption{Additional qualitative results on relational and multi-object prompts with the SDXL backbone.
        Results include unsteered generation, CountGen, Counting Guidance (\emph{Guidance}), and the three \method{} variants.
        Green borders indicate correct object counts, while red borders denote counting errors.
        Prior baselines frequently fail to enforce correct counts or introduce visual artifacts, whereas \method{} improves count fidelity while preserving scene structure and visual coherence, with the adaptive variant yielding the most consistent corrections.
    }
    \label{fig:examples_sdxl}
\end{figure*}

\end{document}